\documentclass{article}

\usepackage{times}
\usepackage{epsfig}
\usepackage{graphicx}
\usepackage{amsmath}
\usepackage{amssymb}
\usepackage{multirow}
\usepackage{algorithm}
\usepackage{algorithmic}
\usepackage{url}

\newcounter{counter_lemma}

\newtheorem{unique}[counter_lemma]{Lemma}
\newtheorem{sign}[counter_lemma]{Lemma}
\newtheorem{hf}[counter_lemma]{Lemma}
\newtheorem{omegaf}[counter_lemma]{Lemma}
\newtheorem{lessThanZero}[counter_lemma]{Lemma}

\newcounter{counter_corolary}
\newtheorem{hsf}[counter_corolary]{Corolary}
\newtheorem{corolarylow}[counter_corolary]{Corolary}

\newcounter{counter_definition}
\newtheorem{aux1}[counter_definition]{Definition}
\newtheorem{aux2}[counter_definition]{Definition}
\newtheorem{aux3}[counter_definition]{Definition}

\newcounter{counter_theorem}

\newtheorem{oneAndInf}[counter_theorem]{Theorem}
\newtheorem{zeroSolution}[counter_theorem]{Theorem}
\newtheorem{singleZF}[counter_theorem]{Theorem}

\newtheorem{remark}{Remark}

\title{Efficient $\ell_1/\ell_{\lowercase {q}}$ Norm Regularization}

\author{Jun Liu \and Jieping Ye}

\date{September 23, 2010}

\begin{document}

\maketitle

\begin{abstract}
Sparse learning has recently received increasing attention in many
areas including machine learning, statistics, and applied
mathematics. The mixed-norm regularization based on the
$\ell_1/\ell_q$ norm with $q>1$ is attractive in many applications of
regression and classification in that it facilitates group sparsity
in the model. The resulting optimization problem is, however,
challenging to solve due to the structure of the
$\ell_1/\ell_q$-regularization. Existing work deals with special
cases including $q=2, \infty$, and they can not be easily extended to
the general case. In this paper, we propose an efficient algorithm
based on the accelerated gradient method for solving the
$\ell_1/\ell_q$-regularized problem, which is applicable for all
values of $q$ larger than $1$, thus significantly extending existing
work. One key building block of the proposed algorithm is the
$\ell_1/\ell_q$-regularized Euclidean projection (EP$_{1q}$). Our
theoretical analysis reveals the key properties of EP$_{1q}$ and
illustrates why EP$_{1q}$ for the general $q$ is significantly more
challenging to solve than the special cases. Based on our theoretical
analysis, we develop an efficient algorithm for EP$_{1q}$ by solving
two zero finding problems. Experimental results demonstrate the
efficiency of the proposed algorithm.
\end{abstract}


\section{Introduction}

Regularization has played a central role in many machine learning
algorithms. The $\ell_1$-regularization has recently received
increasing attention, due to its sparsity-inducing property,
convenient convexity, strong theoretical guarantees, and great
empirical success in various applications. A well-known application
of the $\ell_1$-regularization is the
Lasso~\cite{TIBSHIRANI:1996:ID24}. Recent studies in areas such as
machine learning, statistics, and applied mathematics have witnessed
growing interests in extending the $\ell_1$-regularization to the
$\ell_1/\ell_q$-regularization~\cite{Bach:2008,Duchi:2009:FOLO,Kowalski:2009,Negahban:2009,Obozinski:2007,Yuan:2006,Zhao:2009}.
This leads to the following $\ell_1/\ell_q$-regularized minimization
problem:
\begin{equation}\label{eq:optimization:problem}
    \min_{\mathbf W \in \mathbb{R}^p}  f(\mathbf W) \equiv l(\mathbf W) + \lambda \varpi(\mathbf W),
\end{equation}
where $\mathbf W \in \mathbb{R}^p$ denotes the model parameters,
$l(\cdot)$ is a convex loss dependent on the training samples and
their corresponding responses, $\mathbf W=[\mathbf w_1^{\rm T},
\mathbf w_2^{\rm T}, \ldots, \mathbf w_s^{\rm T}]^{\rm T}$ is divided
into $s$ non-overlapping groups, $\mathbf w_i \in \mathbb{R}^{p_i},
i=1, 2. \ldots, s$, $\lambda >0$ is the regularization parameter, and
\begin{equation}\label{eq:varpi}
     \varpi(\mathbf W)=\sum_{i=1}^s \|\mathbf w_i\|_q
\end{equation}
is the $\ell_1/\ell_q$ norm with $\|\cdot\|_q$ denoting the vector
$\ell_q$ norm ($q\geq 1$). The $\ell_1/\ell_q$-regularization belongs
to the composite absolute penalties (CAP)~\cite{Zhao:2009} family.
When $q=1$, the problem (\ref{eq:optimization:problem}) reduces to
the $\ell_1$-regularized problem. When $q>1$, the
$\ell_1/\ell_q$-regularization facilitates group sparsity in the
resulting model, which is desirable in many applications of
regression and classification.

The practical challenge in the use of the
$\ell_1/\ell_q$-regularization lies in the development of efficient
algorithms for solving (\ref{eq:optimization:problem}), due to the
non-smoothness of the $\ell_1/\ell_q$-regularization. According to
the black-box Complexity
Theory~\cite{NEMIROVSKI:1994:ID6,Nesterov:2004}, the optimal
first-order black-box method~\cite{NEMIROVSKI:1994:ID6,Nesterov:2004}
for solving the class of nonsmooth convex problems converges as
$O(\frac{1}{\sqrt{k}})$ ($k$ denotes the number of iterations), which
is slow. Existing algorithms focus on solving the problem
(\ref{eq:optimization:problem}) or its equivalent constrained version
for $q=2, \infty$, and they can not be easily extended to the general
case. In order to systematically study the practical performance of
the $\ell_1/\ell_q$-regularization family, it is of great importance
to develop efficient algorithms for solving
(\ref{eq:optimization:problem}) for any $q$ larger than $1$.

\subsection{First-Order Methods Applicable for (\ref{eq:optimization:problem})}

When treating $f(\cdot)$ as the general non-smooth convex function,
we can apply the subgradient
descent~\cite{Boyd:2003,NEMIROVSKI:1994:ID6,Nesterov:2004}:
\begin{equation}\label{eq:sd:iteration}
    \mathbf X_{i+1}=\mathbf X_i - \gamma_i \mathbf G_i,
\end{equation}
where $\mathbf G_i \in \partial f(\mathbf X_i)$ is a subgradient of
$f(\cdot)$ at $\mathbf X_i$, and $\gamma_i$ a step size. There are
several different types of step size rules, and more details can be
found in~\cite{Boyd:2003,NEMIROVSKI:1994:ID6}. Subgradient descent is
proven to converge, and it can yield a convergence rate of
$O(1/\sqrt{k})$ for $k$ iterations. However, SD has the following two
disadvantages: 1) SD converges slowly; and 2) the iterates of SD are
very rarely at the points of
non-differentiability~\cite{Duchi:2009:FOLO}, thus it might not
achieve the desirable sparse solution (which is usually at the point
of non-differentiability) within a limited number of iterations.


Coordinate Descent~\cite{Tseng:2001} and its recent
extension---Coordinate Gradient Descent (CGD) can be applied for
optimizing the non-differentiable composite
function~\cite{Tseng:2009:cd}. Coordinate descent has been applied
for the $\ell_1$-norm regularized least
squares~\cite{FRIEDMAN:2008:ID23}, $\ell_1/\ell_{\infty}$-norm
regularized least squares~\cite{Liu:han:2009:blockwise:cd}, and the
sparse group Lasso~\cite{Friedman:2010}. Coordinate gradient descent
has been applied for the group Lasso logistic
regression~\cite{Meier:2008}.  Convergence results for CD and CGD
have been established, when the non-differentiable part is
separable~\cite{Tseng:2001,Tseng:2009:cd}. However, there is no
global convergence rate for CD and CGD (Note, CGD is reported to have
a \emph{local} linear convergence rate under certain
conditions~\cite[Theorem~4]{Tseng:2009:cd}). In addition, it is not
clear whether CD and CGD are applicable for solving the
problem~\eqref{eq:optimization:problem} with an arbitrary $q \ge 1$.

Fixed Point Continuation~\cite{HALE:2007:ID23,Shi:2008} was recently
proposed for solving the $\ell_1$-norm regularized optimization
(i.e., $\varpi(\mathbf W) =\|\mathbf W\|_1$). It is based on the
following fixed point iteration:
\begin{equation}\label{eq:fixed:point}
    \mathbf X_{i+1}=\mathcal {P}^{\varpi}_{\lambda \tau}( \mathbf X_i - \tau l'(\mathbf
    X_i)),
\end{equation}
where $\mathcal {P}^{\varpi}_{\lambda \tau}(\mathbf
W)=\mbox{sgn}(\mathbf W) \odot \max( \mathbf W - \lambda \tau, 0)$
is an operator and $\tau >0$ is the step size. The fixed point
iteration~\eqref{eq:fixed:point} can be applied to
solve~\eqref{eq:optimization:problem} for any convex penalty
$\varpi(\mathbf W)$, with the operator $\mathcal
{P}^{\varpi}_{\lambda \tau}(\cdot)$ being defined as:
\begin{equation}\label{eq:P:definition}
    \mathcal {P}^{\varpi}_{\lambda \tau}(\mathbf W)=\arg \min_{\mathbf X}
    \frac{1}{2}\|\mathbf X- \mathbf W\|_2^2 + \lambda \tau
    \varphi(\mathbf X).
\end{equation}
The operator $\mathcal {P}^{\varpi}_{\lambda \tau}(\cdot)$ is called
the proximal
operator~\cite{Hiriart-Urruty:1993,Moreau:1965,Yosida:1964}, and is
guaranteed to be non-expansive. With a properly chosen $\tau$, the
fixed point iteration \eqref{eq:fixed:point} can converge to the
fixed point $\mathbf X^*$ satisfying
\begin{equation}\label{eq:fixed:point:optimal}
    \mathbf X^*=\mathcal {P}^{\varpi}_{\lambda \tau}( \mathbf X^* - \tau l'(\mathbf
    X^*)).
\end{equation}
It follows from \eqref{eq:P:definition} and
\eqref{eq:fixed:point:optimal} that,
\begin{equation}\label{eq:fixed:optimality:condition}
    \mathbf 0 \in \mathbf X^* - (\mathbf X^* - \tau l'(\mathbf X^*))
    + \lambda \tau \partial \varpi(\mathbf X^*),
\end{equation}
which together with $\tau >0$ indicates that $\mathbf X^*$ is the
optimal solution to~\eqref{eq:optimization:problem}.
In~\cite{Beck:2009,Nesterov:2007}, the gradient descent method is
extended to optimize the composite function in the form
of~\eqref{eq:optimization:problem}, and the iteration step is similar
to~\eqref{eq:fixed:point}. The extended gradient descent method is
proven to yield the convergence rate of $O(1/k)$ for $k$ iterations.
However, as pointed out in~\cite{Beck:2009,Nesterov:2007}, the scheme
in~\eqref{eq:fixed:point} can be further accelerated for
solving~\eqref{eq:optimization:problem}.

Finally, there are various online learning algorithms that have been
developed for dealing with large-scale data, e.g., the truncated
gradient method~\cite{Langford:2009:online}, the forward-looking
subgradient~\cite{Duchi:2009:FOLO}, and the regularized dual
averaging~\cite{Xiao:average:2009} (which is based on the dual
averaging method proposed in~\cite{Nesterov:dual:averaging:2009}).
When applying the aforementioned online learning methods for
solving~\eqref{eq:optimization:problem}, a key building block is the
operator $\mathcal {P}^{\varpi}_{\lambda \tau}(\cdot)$.

\subsection{Main Contributions}

In this paper, we develop an efficient algorithm for solving the
$\ell_1/\ell_q$-regularized problem (\ref{eq:optimization:problem}),
for any $q \geq 1$. 
More specifically, we develop the GLEP$_{1q}$
algorithm\footnote{GLEP$_{1q}$ stands for \textbf{G}roup Sparsity
\textbf{L}earning via the \textbf{$\ell_1/\ell_q$}-regularized
\textbf{E}uclidean \textbf{P}rojection.}, which makes use of the
accelerated gradient method~\cite{Beck:2009,Nesterov:2007} for
minimizing the composite objective functions. GLEP$_{1q}$ has the
following two favorable properties: (1) It is applicable to any
smooth convex loss $l(\cdot)$ (e.g., the least squares loss and the
logistic loss) and any $q\geq 1$. Existing algorithms
are mainly focused on $\ell_1/\ell_2$-regularization and/or
$\ell_1/\ell_{\infty}$-regularization. To the best of our knowledge,
this is the first work that provides an efficient algorithm for
solving (\ref{eq:optimization:problem}) with any $q \geq 1$; and (2)
It achieves a global convergence rate of $O(\frac{1}{k^2})$ ($k$
denotes the number of iterations) for the smooth convex loss
$l(\cdot)$. In comparison, although the methods proposed
in~\cite{Argyriou:2008,DUCHI:2009:icml,Liu:han:2009:blockwise:cd,Obozinski:2007}
converge, there is no known convergence rate; and the method proposed
in~\cite{Meier:2008} has a \emph{local} linear convergence rate under
certain conditions~\cite[Theorem~4]{Tseng:2009:cd}. In addition,
these methods are not applicable for an arbitrary $q \ge 1$.

The main technical contribution of this paper is the development of
an efficient algorithm for computing the $\ell_1/\ell_q$-regularized
Euclidean projection (EP$_{1q}$), which is a key building block in
the proposed GLEP$_{1q}$ algorithm. More specifically, we analyze the
key theoretical properties of the solution of EP$_{1q}$,  based on
which we develop an efficient algorithm for EP$_{1q}$ by solving two
zero finding problems. In addition, our theoretical analysis reveals
why EP$_{1q}$ for the general $q$ is significantly more challenging
than the special cases such as $q=2$. We have conducted experimental
studies to demonstrate the efficiency of the proposed algorithm.

\subsection{Related Work}

We briefly review recent studies on $\ell_1/\ell_q$-regularization,
most of which focus on $\ell_1/\ell_2$-regularization and/or
$\ell_1/\ell_{\infty}$-regularization.


$\ell_1/\ell_2$-Regularization:  The group Lasso was proposed
in~\cite{Yuan:2006} to select the groups of variables for prediction
in the least squares regression. In~\cite{Meier:2008}, the idea of
group lasso was extended for classification by the logistic
regression model, and an algorithm via the coordinate gradient
descent~\cite{Tseng:2009:cd} was developed. In~\cite{Obozinski:2007},
the authors considered joint covariate selection for grouped
classification by the logistic loss, and developed a blockwise
boosting Lasso algorithm with the boosted Lasso~\cite{Zhao:2004}.
In~\cite{Argyriou:2008}, the authors proposed to learn the sparse
representations shared across multiple tasks, and designed an
alternating algorithm. The Spectral projected-gradient (Spg)
algorithm was proposed for solving the $\ell_1/\ell_2$-ball
constrained smooth optimization problem~\cite{BergFriedlander:2008},
equipped with an efficient Euclidean projection that has expected
linear runtime. The $\ell_1/\ell_2$-regularized multi-task learning
was proposed in~\cite{Liu:2009:uai}, and the equivalent smooth
reformulations were solved by the Nesterov's
method~\cite{Nesterov:2004}. 


$\ell_1/\ell_{\infty}$-Regularization: A blockwise coordinate descent
algorithm~\cite{Tseng:2001} was developed for the mutli-task
Lasso~\cite{Liu:han:2009:blockwise:cd}. It was applied to the neural
semantic basis discovery problem. In~\cite{Quattoni:2009}, the
authors considered the multi-task learning via the
$\ell_1/\ell_{\infty}$-regularization, and proposed to solve the
equivalent $\ell_1/\ell_{\infty}$-ball constrained problem by the
projected gradient descent. In~\cite{Negahban:2008:nips}, the authors
considered the multivariate regression via the
$\ell_1/\ell_{\infty}$-regularization, showed that the
high-dimensional scaling of $\ell_1/\ell_{\infty}$-regularization is
qualitatively similar to that of ordinary $\ell_1$-regularization,
and revealed that, when the overlap parameter is large enough
($>2/3$), $\ell_1/\ell_{\infty}$-regularization yields the improved
statistical efficiency over $\ell_1$-regularization.


$\ell_1/\ell_q$-Regularization: In~\cite{DUCHI:2009:icml}, the
authors studied the problem of boosting with structural sparsity, and
developed several boosting algorithms for regularization penalties
including $\ell_1$, $\ell_{\infty}$, $\ell_1/\ell_2$, and
$\ell_1/\ell_{\infty}$. In~\cite{Zhao:2009}, the composite absolute
penalties (CAP) family was introduced, and an algorithm called iCAP
was developed. iCAP employed the least squares loss and the
$\ell_1/\ell_{\infty}$ regularization, and was implemented by the
boosted Lasso~\cite{Zhao:2004}. The multivariate regression with the
$\ell_1/\ell_q$-regularization was studied
in~\cite{Liu:han:2009:report}. In~\cite{Negahban:2009}, a unified
framework was provided for establishing consistency and convergence
rates for the regularized $M$-estimators, and the results for
$\ell_1/\ell_q$ regularization was established.

\subsection{Notation}

Throughout this paper, scalars are denoted by italic letters, and
vectors by bold face letters. Let $\mathbf X, \mathbf Y, \ldots$
denote the $p$-dimensional parameters, $\mathbf x_i, \mathbf y_i,
\ldots$ the $p_i$-dimensional parameters of the $i$-th group, and
$x_i$ the $i$-th component of $\mathbf x$. We denote $\bar q=
\frac{q}{q-1}$, and thus $q$ and $\bar q$ satisfy the following
relationship: $\frac{1}{\bar q} +\frac{1}{q}=1$. We use the following
componentwise operators: $\odot$, $|\cdot|$ and $ {\rm sgn}(\cdot)$.
Specifically, $\mathbf z=\mathbf x \odot \mathbf y$ denotes $z_i=x_i
y_i$; $\mathbf y=|\mathbf x|$ denotes $y_i=|x_i|$; and $\mathbf y=
{\rm sgn}(\mathbf x)$ denotes $y_i={\rm sgn} (x_i)$, where ${\rm sgn}
(\cdot)$ is the signum function: ${\rm sgn} (t)=1$ if $t>0$; ${\rm
sgn} (t)=0$ if $t=0$; and ${\rm sgn} (t)=-1$ if $t<0$.

\section{The Proposed GLEP$_{1q}$ Algorithm}
\label{s:agmeep}


In this section, we present the proposed GLEP$_{1q}$ algorithm for
solving~\eqref{eq:optimization:problem} in the batch learning
setting. The main technical contribution lies in the development of
an efficient algorithm for the $\ell_1/\ell_q$-regularized Euclidean
projection. Specifically, we analyze the key theoretical properties
of the projection in Section~\ref{subsec:property-projection}, and
show that the projection can be computed by solving two zero finding
problems in Section~\ref{subsec:projection-zerofinding}. Note that,
one can develop the online learning algorithm
for~\eqref{eq:optimization:problem} using the online learning
algorithms discussed in the last section, where the
$\ell_1/\ell_q$-regularized Euclidean projection is also a key
building block.

We first construct the following model for approximating the
composite function $\mathcal {M}(\cdot)$ at the point $\mathbf
X$~\cite{Beck:2009,Nesterov:2007}:
\begin{equation}\label{eq:model:mL}
\begin{aligned}
    \mathcal {M}_{L, \mathbf X} (\mathbf Y)  =  [\mbox{loss}(\mathbf X) + \langle \mbox{loss}'(\mathbf X), \mathbf Y - \mathbf X \rangle] + \lambda \varpi(\mathbf Y)+ \frac{L}{2} \|\mathbf Y-\mathbf X\|_2^2,
\end{aligned}
\end{equation}
where $L >0$. In the model $\mathcal {M}_{L, \mathbf X} (\mathbf
Y)$, we apply the first-order Taylor expansion at the point $\mathbf
X$ (including all terms in the square bracket) for the smooth loss
function $l(\cdot)$, and directly put the non-smooth penalty
$\varpi(\cdot)$ into the model. The regularization term $\frac{L}{2}
\|\mathbf Y-\mathbf X\|_2^2$ prevents $\mathbf Y$ from walking far
away from $\mathbf X$, thus the model can be a good approximation to
$f(\mathbf Y)$ in the neighborhood of $\mathbf X$.

The accelerated gradient method is based on two sequences $\{\mathbf
X_i\}$ and $\{\mathbf S_i \}$ in which $\{\mathbf X_i\}$ is the
sequence of approximate solutions, and $\{\mathbf S_i \}$ is the
sequence of search points. The search point $\mathbf S_i$ is the
affine combination of $\mathbf X_{i-1}$ and $\mathbf X_i$ as
\begin{equation}\label{eq:sk}
    \mathbf S_i= \mathbf X_i + \beta_i (\mathbf X_i - \mathbf
    X_{i-1}),
\end{equation}
where $\beta_i$ is a properly chosen coefficient. The approximate
solution $\mathbf X_{i+1}$ is computed as the minimizer of $\mathcal
{M}_{L_i, \mathbf S_i} (\mathbf Y)$:
\begin{equation}\label{eq:xkplus1}
    \mathbf X_{i+1}= \arg \min_{\mathbf Y} \mathcal {M}_{L_i, \mathbf S_i} (\mathbf   Y),
\end{equation}
where $L_i$ is determined by line search, e.g., the Armijo-Goldstein
rule so that $L_i$ should be appropriate for $\mathbf S_i$.

\begin{algorithm}
  \caption{GLEP$_{1q}$: \textbf{G}roup Sparsity \textbf{L}earning via the
\textbf{$\ell_1/\ell_q$}-regularized \textbf{E}uclidean
\textbf{P}rojection}
  \label{algorithm:GLEP$_{1q}$}
\begin{algorithmic}[1]
  \REQUIRE $\lambda_1 \geq 0, \lambda_2 \geq 0,  L_0 >0, \mathbf X_0, k$
  \ENSURE  $\mathbf X_{k+1}$
    \STATE Initialize $\mathbf X_1=\mathbf X_0$, $\alpha_{-1}=0$, $\alpha_0=1$, and $L=L_0$.
    \FOR{$i=1$ to $k$}
     \STATE Set $\beta_i= \frac{\alpha_{i-2}-1}{\alpha_{i-1}}$, $\mathbf S_i = \mathbf X_i + \beta_i (\mathbf X_i  - \mathbf X_{i-1})$
     \STATE Find the smallest $L=L_{i-1}, 2L_{i-1}, \ldots $ such that $$f( \mathbf X_{i+1}) \leq \mathcal {M}_{L,\mathbf S_i}( \mathbf X_{i+1} ),$$
     where $\mathbf X_{i+1}= \arg \min_{\mathbf Y} \mathcal {M}_{L, \mathbf S_i} (\mathbf  Y)$
    \STATE Set $L_i=L$ and $\alpha_{i+1}=\frac{1+\sqrt{1+4  \alpha_i^2}}{2}$
    \ENDFOR
\end{algorithmic}
\end{algorithm}

The algorithm for solving~\eqref{eq:optimization:problem} is
presented in Algorithm~\ref{algorithm:GLEP$_{1q}$}. GLEP$_{1q}$
inherits the optimal convergence rate of $O(1/k^2)$ from the
accelerated gradient method. In
Algorithm~\ref{algorithm:GLEP$_{1q}$}, a key subroutine
is~\eqref{eq:xkplus1}, which can be computed as $\mathbf
X_{i+1}=\pi_{1q}(\mathbf S_i- l'(\mathbf S_i) /L_i, \lambda / L_i)$,
where $\pi_{1q}(\cdot)$ is the $\ell_1/\ell_q$-regularized Euclidean
projection (EP$_{1q}$) problem:
\begin{equation}\label{eq:projection:1q}
    \pi_{1q}(\mathbf V, \lambda)= \arg \min_{\mathbf X \in \mathbb{R}^p} \frac{1}{2} \|\mathbf X-\mathbf V\|_2^2 +  \lambda \sum_{i=1}^s \|\mathbf
    x_i\|_q.
\end{equation}
The efficient computation of~\eqref{eq:projection:1q} for any $q > 1$
is the main technical contribution of this paper.
%
Note that the $s$ groups in \eqref{eq:projection:1q} are independent.
Thus the optimization in (\ref{eq:projection:1q}) decouples into a
set of $s$ independent $\ell_q$-regularized Euclidean projection
problems:
\begin{equation}\label{eq:subproblem}
    \pi_q (\mathbf v)= \arg \min_{\mathbf x \in \mathbb{R}^n} \left(g(\mathbf x)= \frac{1}{2} \|\mathbf x -\mathbf v\|_2^2 + \lambda
    \|\mathbf x\|_q\right),
\end{equation}
where $n=p_i$ for the $i$-th group. Next, we study the key properties
of (\ref{eq:subproblem}).

\subsection{Properties of the Optimal Solution to (\ref{eq:subproblem})}
\label{subsec:property-projection}

The function $g(\cdot)$ is strictly convex, and thus it has a unique
minimizer, as summarized below:
\begin{unique}\label{lemma:unique}
The problem (\ref{eq:subproblem}) has a unique minimizer.
\end{unique}

Next, we show that the optimal solution to (\ref{eq:subproblem}) is
given by zero under a certain condition, as summarized in the
following theorem:
\begin{zeroSolution}\label{theorem:solution}
$\pi_q (\mathbf v) = \mathbf 0$ if and only if $\lambda \geq
\|\mathbf v\|_{\bar q}$.
\end{zeroSolution}

\noindent \textbf{Proof: } Let us first compute the directional
derivative of $g(\mathbf x)$ at the point $\mathbf 0$:
\begin{equation*}
    D g(\mathbf 0)[\mathbf u] = \lim_{ \alpha \downarrow 0 }
    \frac{1}{\alpha}[g(\alpha \mathbf u) - g(\mathbf 0)]
                              = -\langle \mathbf v, \mathbf u \rangle +
                              \lambda \|\mathbf u\|_q,
\end{equation*}
where $\mathbf u$ is a given direction. According to the
H\"{o}lder's inequality, we have
\begin{equation*}
    |\langle \mathbf u, \mathbf v \rangle | \leq \|\mathbf u\|_q
    \|\mathbf v\|_{\bar q}, \forall \mathbf u.
\end{equation*}
Therefore, we have
\begin{equation}\label{eq:direction:zero}
    D g(\mathbf 0)[\mathbf u] \geq 0, \forall \mathbf u,
\end{equation}
if and only if $\lambda \geq \|\mathbf v\|_{\bar q}$. The result
follows, since (\ref{eq:direction:zero}) is the necessary and
sufficient condition for $\mathbf 0$ to be the optimal solution of
(\ref{eq:subproblem}). \hfill $\Box$

Next, we focus on solving (\ref{eq:subproblem}) for $0<\lambda <
\|\mathbf v\|_{\bar q}$. We first consider solving
(\ref{eq:subproblem}) in the case of $1 < q < \infty$, which is the
main technical contribution of this paper. We begin with a lemma
that summarizes the key properties of the optimal solution to the
problem (\ref{eq:subproblem}):

\begin{sign}\label{lemma:sign}
Let $1<q<\infty$ and $0<\lambda < \|\mathbf v\|_{\bar q} $. Then,
$\mathbf x^*$ is the optimal solution to the problem
(\ref{eq:subproblem}) if and if only it satisfies:
\begin{equation}\label{eq:x:nonzero}
    \mathbf x^* + \lambda \|\mathbf x^*\|_q^{1-q} {\mathbf
    x^*}^{(q-1)} = \mathbf v,
\end{equation}
where $\mathbf y \equiv \mathbf x^{(q-1)}$ is defined
component-wisely as:
   $ y_i={\rm sgn}(x_i) |x_i|^{q-1}$.
Moreover, we have
\begin{equation}\label{eq:v:absv}
    \pi_q(\mathbf v)= {\rm sgn} (\mathbf v) \odot \pi_q(|\mathbf
    v|),
\end{equation}
\begin{equation}\label{eq:sign:keep}
    {\rm sgn}(\mathbf x^*)={\rm sgn} (\mathbf v),
\end{equation}
\begin{equation}\label{eq:x:bound}
    0 < |x^*_i| < |v_i|, \forall i \in \{i|v_i \neq 0\}.
\end{equation}
\end{sign}

\noindent \textbf{Proof: } Since $\lambda < \|\mathbf v\|_{\bar q}$,
it follows from Theorem~\ref{theorem:solution} that the optimal
solution $\mathbf x^* \neq \mathbf 0$. $\|\mathbf x\|_q$ is
differentiable when $\mathbf x \neq \mathbf 0$, so is $g(\mathbf
x)$. Therefore, the sufficient and necessary condition for $\mathbf
x^*$ to be the solution of (\ref{eq:subproblem}) is $g'(\mathbf
x^*)=0$, i.e., (\ref{eq:x:nonzero}). Denote $c^* \equiv \lambda
\|\mathbf x^*\|_q^{1-q} >0$. It follows from (\ref{eq:x:nonzero})
that (\ref{eq:v:absv}) holds, and
\begin{equation}\label{eq:x:comp}
    \mbox{sgn}(x^*_i) \left( |x^*_i| + c^*  |x^*_i|^{q-1} \right)=v_i,
\end{equation}
from which we can verify (\ref{eq:sign:keep}) and
(\ref{eq:x:bound}). \hfill $\Box$

\begin{figure}
  \centering
  \includegraphics[width=2.1in]{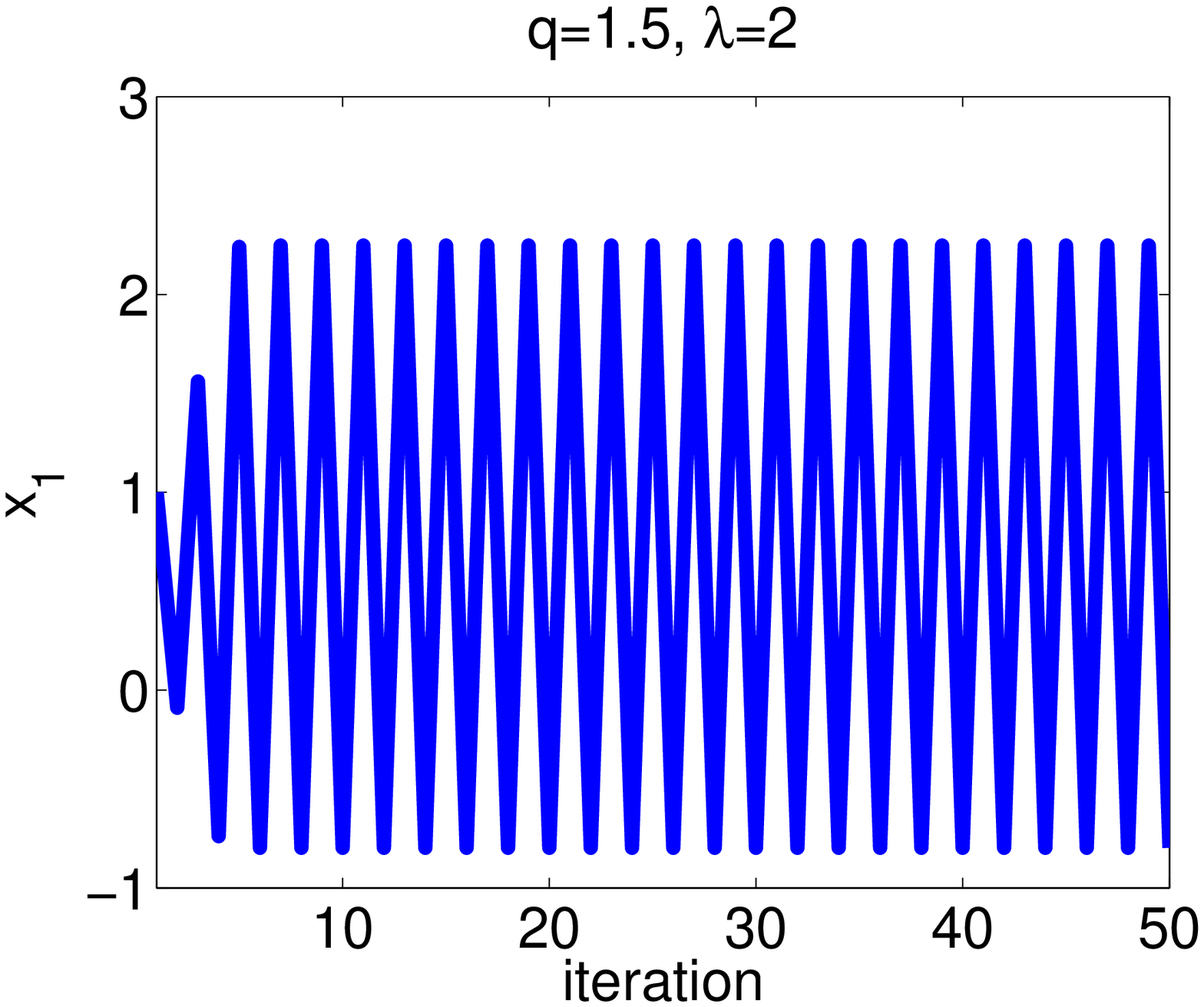} \hspace{0.1in}
  \includegraphics[width=2.1in]{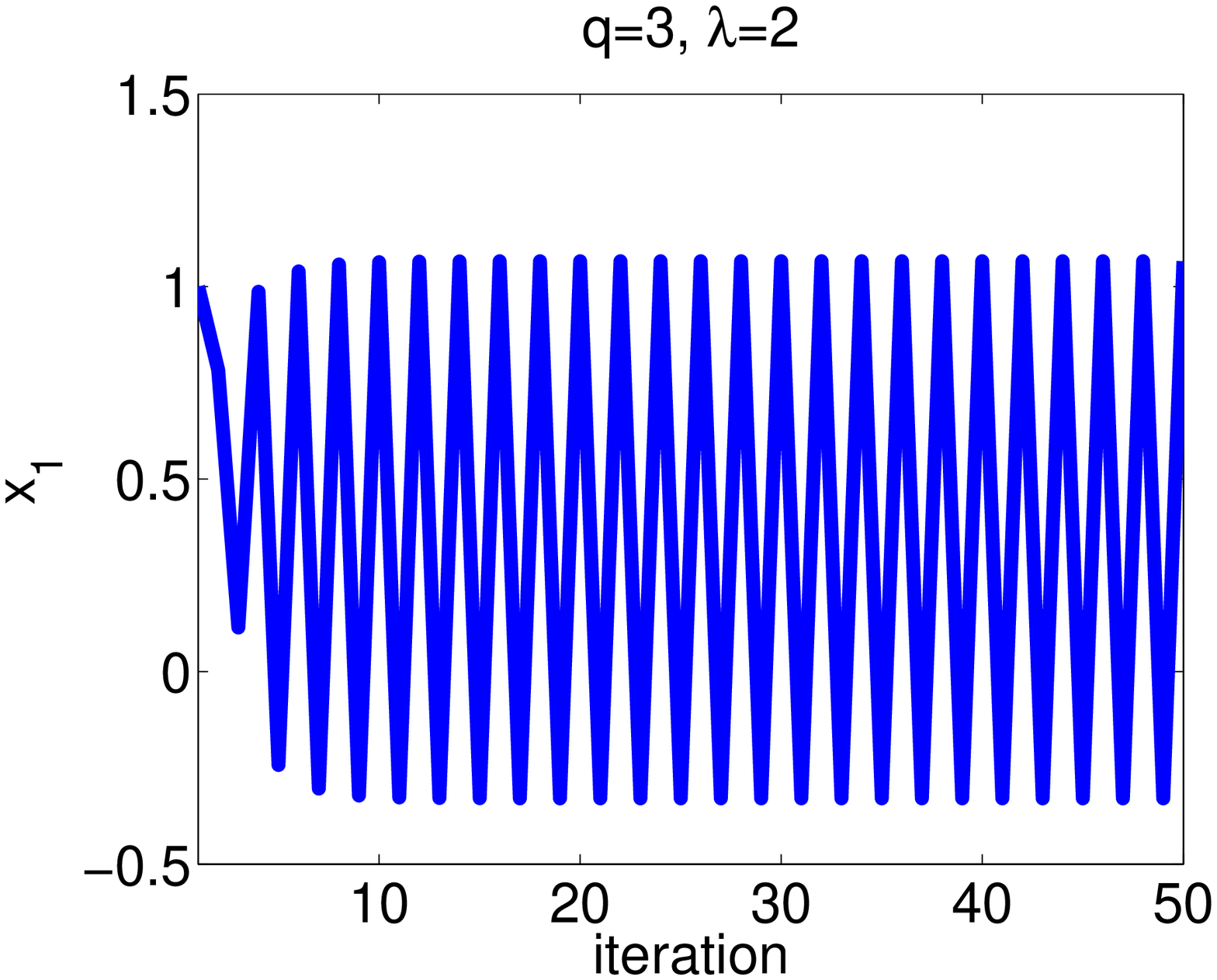}\\[-0.25cm]
  \caption{{\small Illustration of the failure of the fixed point iteration $\mathbf x
= \mathbf v - \lambda \|\mathbf x\|_q^{1-q} {\mathbf x}^{(q-1)}$ for
solving (\ref{eq:subproblem}). We set $\mathbf v=[1, 3]^{\rm T}$ and
the starting point $\mathbf x=[1, 3]^{\rm T}$. The vertical axis
denotes the values of $x_1$ during the
iterations.}}\label{fig:fp:fail}
\end{figure}

It follows from Lemma~\ref{lemma:sign} that i) if $v_i=0$ then
$x_i^*=0$; and ii) $\pi_q(\mathbf v)$ can be easily obtained from
$\pi_q(|\mathbf v|)$. Thus, we can restrict our following discussion
to $\mathbf v> \mathbf 0$, i.e., $v_i>0, \forall i$. It is clear
that, the analysis can be easily extended to the general $\mathbf v$.
The optimality condition in (\ref{eq:x:nonzero}) indicates that
$\mathbf x^*$ might be solved via the fixed point iteration
$$\mathbf x =\eta(\mathbf x) \equiv \mathbf v - \lambda \|\mathbf x\|_q^{1-q}
{\mathbf x}^{(q-1)},$$ which is, however, not guaranteed to converge
(see Figure~\ref{fig:fp:fail} for examples), as $\eta(\cdot)$ is not
necessarily a contraction
mapping~\cite[Proposition~3]{Kowalski:2009}. In addition, $\mathbf
x^*$ cannot be trivially solved by firstly guessing $c=\|\mathbf
x\|_q^{1-q}$ and then finding the root of $\mathbf x + \lambda c
{\mathbf x}^{(q-1)} = \mathbf v$, as when $c$ increases, the values
of $\mathbf x$ obtained from $\mathbf x + \lambda c {\mathbf
x}^{(q-1)} = \mathbf v$ decrease, so that $c=\|\mathbf x\|_q^{1-q}$
increases as well (note that $1-q<0$).

\subsection{Computing the Optimal Solution $\mathbf x^*$ by Zero Finding}
\label{subsec:projection-zerofinding}

In the following, we show that $\mathbf x^*$ can be obtained by
solving two zero finding problems. Below, we construct our first
auxiliary function $h_c^v(\cdot)$ and reveal its properties:

\begin{aux3}[Auxiliary Function $h_c^v(\cdot)$ ]\label{def:auxiliary-1}
Let $c>0$, $1 < q < \infty$, and $v >0$. We define the auxiliary
function $h_c^v(\cdot)$ as follows:
\begin{equation}\label{eq:h}
    h_c^v(x)= x + c x^{q-1} - v, 0\leq x \leq v.
\end{equation}
\end{aux3}

\begin{hf}\label{lemma:hf}
Let $c>0$, $1 < q < \infty$, and $v >0$. Then, $h_c^v(\cdot)$ has a
unique root in the interval $(0, v)$.
\end{hf}

\noindent \textbf{Proof: } It is clear that $h_c^v(\cdot)$ is
continuous and strictly increasing in the interval $[0, v]$,
$h_c^v(0)=-v<0$, and $h_c^v(v)= c v^{q-1}
>0$. According to the Intermediate Value Theorem, $h_c^v(\cdot)$ has a
unique root lying in the interval $(0, v)$. This concludes the
proof. \hfill $\Box$

\begin{hsf}\label{corolary:hsf}
Let $\mathbf x, \mathbf v \in \mathbb{R}^n$, $c>0$, $1 < p <
\infty$, and $\mathbf v >\mathbf 0$. Then, the function
\begin{equation}\label{eq:simple:problem}
    \varphi_c^{\mathbf v}(\mathbf x)=\mathbf x + c \mathbf x^{(q-1)}
    -\mathbf v, \mathbf 0 < \mathbf x
    < \mathbf v
\end{equation}
has a unique root.
\end{hsf}

Let $\mathbf x^*$ be the optimal solution satisfying
(\ref{eq:x:nonzero}). Denote $c^* = \lambda \|\mathbf x^*\|_q^{1-q}$.
It follows from Lemma~\ref{lemma:sign} and
Corollary~\ref{corolary:hsf} that $\mathbf x^*$ is the unique root of
$\varphi_{c^*}^{\mathbf v} (\cdot)$ defined in
(\ref{eq:simple:problem}), provided that the optimal $c^*$ is known.
Our methodology for computing $\mathbf x^*$ is to first compute the
optimal $c^*$ and then compute $\mathbf x^*$ by computing the root of
$\varphi_{c^*}^{\mathbf v}(\cdot)$. Next, we show how to compute the
optimal $c^*$ by solving a single variable zero finding problem. We
need our second auxiliary function $\omega(\cdot)$ defined as
follows:

\begin{aux2}[Auxiliary Function $\omega(\cdot)$] \label{def:auxiliary-2}
Let $1 < q < \infty $ and $v >0$. We define the auxiliary function
$\omega(\cdot)$ as follows:
\begin{equation}\label{eq:omega}
    c=\omega(x)= (v-x) /  x^{q-1}, 0< x \leq v.
\end{equation}
\end{aux2}

\begin{omegaf}\label{lemma:omegaf}
In the interval $(0, v]$, $c=\omega(x)$ is i) continuously
differentiable, ii) strictly decreasing, and iii) invertible.
Moreover, in the domain $[0, \infty)$, the inverse function
$x=\omega^{-1}(c)$ is continuously differentiable and strictly
decreasing.
\end{omegaf}
\noindent \textbf{Proof: } It is easy to verify that, in the
interval $(0, v]$, $c=\omega(x)$ is continuously differentiable with
a non-positive gradient, i.e., $\omega'(x) < 0$. Therefore, the
results follow from the Inverse Function Theorem. \hfill $\Box$

It follows from Lemma~\ref{lemma:omegaf} that given the optimal
$c^*$ and $\mathbf v$, the optimal $\mathbf x^*$ can be computed via
the inverse function $\omega^{-1}(\cdot)$, i.e., we can represent
$\mathbf x^*$ as a function of $c^*$. Since $\lambda \|\mathbf
x^*\|_q^{1-q}-c^* =0$ by the definition of $c^*$, the optimal $c^*$
is a root of our third auxiliary function $\phi(\cdot)$ defined as
follows:

\begin{aux1}
[Auxiliary Function $\phi(\cdot)$] \label{def:auxiliary-3} Let $1 < q
< \infty$, $0<\lambda < \|\mathbf v\|_{\bar q}$, and $\mathbf v
>\mathbf 0$. We define the auxiliary function $\phi(\cdot)$ as follows:
\begin{equation}\label{eq:phi:c}
    \phi(c)=\lambda \psi(c)-c, c \geq 0,
\end{equation}
where
\begin{equation}\label{eq:psi:c}
    \psi(c)=\left(\sum_{i=1}^n
    (\omega_i^{-1}(c))^q\right)^{\frac{1-q}{q}},
\end{equation}
and $\omega_i^{-1}(c)$ is the inverse function of
\begin{equation}\label{eq:c:x}
    \omega_i(x)=(v_i-x) / x^{q-1}, 0< x \leq v_i.
\end{equation}
\end{aux1}

Recall that we assume $0<\lambda < \|\mathbf v\|_{\bar q}$
(otherwise the optimal solution is given by zero from
Theorem~\ref{theorem:solution}). The following lemma summarizes the
key properties of the auxiliary function $\phi(\cdot)$:

\begin{lessThanZero}\label{lemma:lessThanZero}
Let $1 < q < \infty$, $0<\lambda < \|\mathbf v\|_{\bar q}$, $\mathbf
v >\mathbf 0$, and
\begin{equation}\label{eq:epsilon}
    \epsilon=(\|\mathbf v\|_{\bar q} - \lambda) / \|\mathbf v\|_{\bar q}.
\end{equation}
Then, $\phi(\cdot)$ is continuously differentiable in the interval
$[0, \infty)$. Moreover, we have $$\phi(0)=\lambda \| \mathbf
v\|_q^{1-q}
>0,\phi(\overline c) \leq 0,$$ where
\begin{equation}\label{eq:c:tilde}
    \overline c= \max_i c_i,
\end{equation}
\begin{equation}\label{eq:ci:vi}
    c_i=\omega_i(v_i \epsilon), i=1, 2, \ldots, n.
\end{equation}
\end{lessThanZero}

 \noindent \textbf{Proof: } From Lemma~\ref{lemma:omegaf}, the function
$\omega_i^{-1}(c)$ is continuously differentiable in $[0, \infty)$.
It is easy to verify that $\omega_i^{-1}(c) >0, \forall c \in [0,
\infty)$. Thus, $\phi(\cdot)$ in~(\ref{eq:phi:c}) is continuously
differentiable in $[0, \infty)$.

It is clear that $\phi(0)=\lambda \|\mathbf v\|_q^{1-q}
>0$. Next, we show $\phi(\overline c) \leq 0$. Since $0<\lambda <
\|\mathbf v\|_{\bar q}$, we have
\begin{equation}\label{epsilon:range}
    0 <\epsilon <1.
\end{equation}
It follows from (\ref{eq:c:x}), (\ref{eq:c:tilde}), (\ref{eq:ci:vi})
and (\ref{epsilon:range}) that $0 < c_i \leq \overline c, \forall
i$. Let $\mathbf x=[x_1, x_2, \ldots, x_n]^{\rm T}$ be the root of
$\varphi_{\overline c}^{\mathbf v} (\cdot)$ (see
Corollary~\ref{corolary:hsf}). Then, $x_i=\omega^{-1}_i(\overline
c)$. Since $\omega^{-1}_i(\cdot)$ is strictly decreasing (see
Lemma~\ref{lemma:omegaf}), $c_i \leq \overline c$, $v_i
\epsilon=\omega_i^{-1}(c_i)$, and $x_i=\omega_i^{-1}(\overline c)$,
we have
\begin{equation}\label{eq:x:upper:bound}
    x_i \leq v_i \epsilon.
\end{equation}
Combining (\ref{eq:c:x}), (\ref{eq:x:upper:bound}), and $\overline c=
\omega_i(x_i)$, we have $ \overline c \geq
v_i(1-\epsilon)/x_i^{q-1}$, since $\omega_i(\cdot)$ is strictly
decreasing. It follows that $x_i \geq
\left(\frac{v_i(1-\epsilon)}{\overline c}\right)^{\frac{1}{q-1}}$.
Thus, the following holds:
\begin{equation*}
    \psi(\overline c)=  \left(\sum_{i=1}^n (\omega_i^{-1}(\overline c))^q
    \right)^{\frac{1-q}{q}} = \left(\sum_{i=1}^n x_i^q \right)^{\frac{1-q}{q}}
    \leq \frac{\overline c}{\|\mathbf v\|_{\bar q} (1-\epsilon)},
\end{equation*}
which leads to
\begin{equation*}\label{eq:phi:tilde:c}
    \phi(\overline c) =\lambda \psi(\overline c) -\overline c \leq \overline c
    \left(\frac{\lambda}{\|\mathbf v\|_{\bar q} (1-\epsilon)}-1 \right)=0,
\end{equation*}
where the last equality follows from (\ref{eq:epsilon}). \hfill
$\Box$

\begin{corolarylow}\label{corolary:lower}
Let $1 < q < \infty$, $0<\lambda < \|\mathbf v\|_{\bar q}$, $\mathbf
v >\mathbf 0$, and $ \underline c= \min_i c_i$, where $c_i$'s are
defined in (\ref{eq:ci:vi}). We have $0<\underline c \leq \overline
c$ and $\phi(\underline c) \geq 0$.
\end{corolarylow}

Following Lemma~\ref{lemma:lessThanZero} and
Corollary~\ref{corolary:lower}, we can find at least one root of
$\phi(\cdot)$ in the interval $[\underline c, \overline c]$. In the
following theorem, we show that $\phi(\cdot)$ has a unique root:

\begin{singleZF} \label{theorem:singleZF}
Let $1 < q < \infty$, $0<\lambda < \|\mathbf v\|_{\bar q}$, and
$\mathbf v >\mathbf 0$. Then, in $[\underline c, \overline c]$,
$\phi(\cdot)$ has a unique root, denoted by $c^*$, and the root of
$\varphi_{c^*}^{\mathbf v}(\cdot)$ is the optimal solution to
(\ref{eq:subproblem}).
\end{singleZF}

 \noindent \textbf{Proof: } From Lemma~\ref{lemma:lessThanZero} and
Corollary~\ref{corolary:lower}, we have $\phi(\overline c) \leq 0$
and $\phi(\underline c) \geq 0$. If either $\phi(\overline c) = 0$
or $\phi(\underline c) = 0$, $\overline c$ or $\underline c$ is a
root of $\phi(\cdot)$. Otherwise, we have $\phi(\underline c)
\phi(\overline c) <0$. As $\phi(\cdot)$ is continuous in $[0,
\infty)$, we conclude that $\phi(\cdot)$ has a root in $(\underline
c, \overline c)$ according to the Intermediate Value Theorem.

Next, we show that $\phi(\cdot)$ has a unique root in the interval
$[0, \infty)$. We prove this by contradiction. Assume that
$\phi(\cdot)$ has two roots: $0< c_1 <c_2$. From
Corollary~\ref{corolary:hsf}, $\varphi_{c_1}^{\mathbf v}(\cdot)$ and
$\varphi_{c_2}^{\mathbf v}(\cdot)$ have unique roots. Denote $\mathbf
x^1=[x_1^1, x_2^1, \ldots, x_n^1]^{\rm T}$ and $\mathbf x^2=[x_1^2,
x_2^2, \ldots, x_n^2]^{\rm T}$ as the roots of
$\varphi_{c_1}^{\mathbf v}(\cdot)$ and $\varphi_{c_2}^{\mathbf
v}(\cdot)$, respectively. We have $0< x_i^1, x_i^2 < v_i, \forall i$.
It follows from (\ref{eq:phi:c}-\ref{eq:c:x}) that
\begin{eqnarray}
   \mathbf x^1 + \lambda \|\mathbf x^1\|_q^{1-q} {\mathbf x^1}^{(q-1)} -
   \mathbf
   v = \mathbf 0, \nonumber \\
   \mathbf x^2 + \lambda \|\mathbf x^2\|_q^{1-q} {\mathbf x^2}^{(q-1)} -
   \mathbf
   v = \mathbf 0. \nonumber
\end{eqnarray}
According to Lemma~\ref{lemma:sign}, $\mathbf x^1$ and $\mathbf x^2$
are the optimal solution of (\ref{eq:subproblem}). From
Lemma~\ref{lemma:unique}, we have $\mathbf x^1=\mathbf x^2$.
However, since $x_i^1=\omega_i^{-1}(c_1)$,
$x_i^2=\omega_i^{-1}(c_2)$, $\omega_i^{-1}(\cdot)$ is a strictly
decreasing function in $[0, \infty)$ by Lemma~\ref{lemma:omegaf},
and $c_1 < c_2$, we have $x_i^1 > x_i^2, \forall i$. This leads to a
contradiction. Therefore, we conclude that $\phi(\cdot)$ has a
unique root in $[\underline c, \overline c]$.

From the above arguments, it is clear that, the root of
$\varphi_{c^*}^{\mathbf v}(\cdot)$ is the optimal solution to
(\ref{eq:subproblem}). \hfill $\Box$

\begin{remark} \label{remark:q=2}
When $q=2$, we have $\underline c=\overline c
=\frac{\lambda}{\|\mathbf v\|_2 -\lambda}$. It is easy to verify
that $\phi(\underline c)=\phi(\overline c)=0$ and
\begin{equation}\label{eq:p:2:solution}
    \pi_2(\mathbf v)=\frac{\|\mathbf v\|_2 - \lambda}{\|\mathbf v\|_2} \mathbf
    v.
\end{equation}
Therefore, when $q=2$, we obtain a closed-form solution.
\end{remark}


\subsection{Solving the Zero Finding Problem by Bisection}

Let $1 < q < \infty$, $0<\lambda < \|\mathbf v\|_{\bar q}$, $\mathbf
v
>\mathbf 0$, $\overline v= \max_i v_i$, $\underline v =\min_i v_i$,
and $\delta
>0$ be a small constant (e.g., $\delta=10^{-8}$ in our experiments).
When $q>2$, we have
$$\underline c =
\frac{1-\epsilon}{\epsilon^{q-1}\overline v^{q-2}} \quad \mbox{  and
} \quad \overline c = \frac{1-\epsilon}{\epsilon^{q-1}\underline
v^{q-2}}.$$ When $1 < q < 2$, we have
$$\underline c =
\frac{1-\epsilon}{\epsilon^{q-1}\underline v^{q-2}} \quad \mbox{ and
} \quad \overline c = \frac{1-\epsilon}{\epsilon^{q-1}\overline
v^{q-2}}.$$ If either $\phi(\overline c)=0$ or $\phi(\underline
c)=0$, $\overline c$ or $\underline c$ is the unique root of
$\phi(\cdot)$. Otherwise, we can find the unique root of
$\phi(\cdot)$ by bisection in the interval $(\underline c, \overline
c)$, which costs at most
$$N=\log_2 \frac{(1-\epsilon)|\overline v^{q-2}- \underline
v^{q-2}|}{\epsilon^{q-1}\overline v^{q-2} \underline v^{q-2}\delta}$$
iterations for achieving an accuracy of $\delta$. Let $[c_1, c_2]$ be
the current interval of uncertainty, and we have computed
$\omega_i^{-1}(c_1)$ and $\omega_i^{-1}(c_2)$ in the previous
bisection iterations. Setting $c=\frac{c_1+c_2}{2}$, we need to
evaluate $\phi(c)$ by computing $\omega_i^{-1}(c), i=1, 2, \ldots,
n$. It is easy to verify that $\omega_i^{-1}(c)$ is the root of
$h_c^{v_i}(\cdot)$ in the interval $(0, v_i)$. Since
$\omega_i^{-1}(\cdot)$ is a strictly decreasing function (see
Lemma~\ref{lemma:omegaf}), the following holds:
$$\omega_i^{-1}(c_2 )
< \omega_i^{-1}(c) < \omega_i^{-1}(c_1),$$ and thus
$\omega_i^{-1}(c)$ can be solved by bisection using at most
$$\log_2 \frac{\omega_i^{-1}(c_2)-\omega_i^{-1}(c_1)}{\delta} < \log_2
\frac{v_i}{\delta} \leq \log_2 \frac{\overline v}{\delta}$$ iterations
for achieving an accuracy of $\delta$. For given $\mathbf v,
\lambda$, and $\delta$, $N$ and $\overline v$ are constant, and thus
it costs $O(n)$ for finding the root of $\phi(\cdot)$. Once $c^*$,
the root of $\phi(\cdot)$ is found, it costs $O(n)$ flops to compute
$\mathbf x^*$ as the unique root of $\varphi_{c^*}^{\mathbf
v}(\cdot)$. Therefore, the overall time complexity for solving
(\ref{eq:subproblem}) is $O(n)$.

We have shown how to solve (\ref{eq:subproblem}) for $1 < q <
\infty$. For $q=1$, the problem (\ref{eq:subproblem}) is reduced to
the one used in the standard Lasso, and it has the following
closed-form solution~\cite{Beck:2009}:
\begin{equation}\label{eq:x:solution}
    \pi_1(\mathbf v)={\rm sgn}(\mathbf v) \odot \max(|\mathbf
    v|-\lambda,0).
\end{equation}
For $q=\infty$,  the problem (\ref{eq:subproblem}) can computed
via~\eqref{eq:x:solution}, as summarized in the following theorem:
\begin{oneAndInf}\label{theorem:oneAndInf}
Let $q=\infty$, $\bar q=1$, and $0<\lambda < \|\mathbf v\|_{\bar
q}$. Then we have
\begin{equation}\label{eq:x:infty}
    \pi_{\infty}(\mathbf v)={\rm sgn}(\mathbf v) \odot \min( |\mathbf
    v|, t^* ),
\end{equation}
where $t^*$ is the unique root of
\begin{equation}\label{eq:zero:finding:l1}
    h(t)=\sum_{i=1}^n \max( |v_i| - t, 0) - \lambda.
\end{equation}
\end{oneAndInf}

\noindent \textbf{Proof: } Making use of the property that
$\|\mathbf x\|_{\infty} = \max_{\|\mathbf y\|_1 \leq 1} \langle
\mathbf y, \mathbf x \rangle$,
we can rewrite (\ref{eq:subproblem}) in the case of $q=\infty$ as
\begin{equation}\label{eq:minmax}
    \min_{\mathbf x} \max_{\mathbf y: \|\mathbf y\|_1 \leq \lambda } s(\mathbf x, \mathbf y) \equiv \frac{1}{2} \|\mathbf x -\mathbf v\|_2^2 + \langle \mathbf y, \mathbf x
    \rangle.
\end{equation}
The function $s(\mathbf x, \mathbf y)$ is continuously
differentiable in both $\mathbf x$ and $\mathbf y$, convex in
$\mathbf x$ and concave in $\mathbf y$, and the feasible domains are
solids. According to the well-known von Neumann
Lemma~\cite{NEMIROVSKI:1994:ID6}, the min-max problem
(\ref{eq:minmax}) has a saddle point, and thus the minimization and
maximization can be exchanged. Setting the derivative of $s(\mathbf
x, \mathbf y)$ with respect to $\mathbf x$ to zero, we have
\begin{equation}\label{eq:x:y}
    \mathbf x = \mathbf v -\mathbf y.
\end{equation}
Thus we obtain the following problem:
\begin{equation}\label{eq:l1:proj}
    \min_{\mathbf y: \|\mathbf y\|_1 \leq \lambda} \frac{1}{2} \|\mathbf y -\mathbf
    v\|_2^2,
\end{equation}
which is the problem of the Euclidean projection onto the $\ell_1$
ball~\cite{BergFriedlander:2008,DUCHI:2009:icml,Liu:2009:icml}. It
has been shown that the optimal solution $\mathbf y^*$to
(\ref{eq:l1:proj}) for $\lambda < \|\mathbf v\|_{1}$ can be obtained
by first computing $t^*$ as the unique root of
(\ref{eq:zero:finding:l1}) in linear time, and then computing
$\mathbf y^*$ as
\begin{equation}\label{eq:y:solution}
    \mathbf y^*={\rm sgn}(\mathbf v) \odot \max( |\mathbf v| - t^*, 0).
\end{equation}
It follows from (\ref{eq:x:y}) and (\ref{eq:y:solution}) that
(\ref{eq:x:infty}) holds. \hfill $\Box$

We conclude this section by summarizing the main steps for solving
the $\ell_q$-regularized Euclidean projection in
Algorithm~\ref{algorithm:Eq}.

\begin{algorithm}
  \caption{Ep$_{q}$: $\ell_q$-regularized Euclidean projection}
  \label{algorithm:Eq}
\begin{algorithmic}[1]
  \REQUIRE $\lambda>0, q \ge 1, \mathbf v \in \mathbb{R}^n$
  \ENSURE  $\mathbf x^*= \pi_q(\mathbf v)=\arg \min_{\mathbf x \in \mathbb{R}^n}  \frac{1}{2} \|\mathbf x -\mathbf v\|_2^2 + \lambda
    \|\mathbf x\|_q$
    \STATE Compute $\bar q= \frac{q}{q-1}$
    \IF{$\|\mathbf v\|_{\bar q} \le \lambda $}
    \STATE Set $\mathbf x^*=\mathbf 0$, return
    \ENDIF
    \IF{$q=1$}
    \STATE Set $\mathbf x^*={\rm sgn}(\mathbf v) \odot \max(|\mathbf
    v|-\lambda,0)$
    \ELSIF{$q=2$}
    \STATE Set $\mathbf x^*=\frac{\|\mathbf v\|_2 - \lambda}{\|\mathbf v\|_2} \mathbf  v$
    \ELSIF{$q=\infty$}
          \STATE Obtain $t^*$, the unique root of $h(t)$, via the improved bisection method~\cite{Liu:2009:icml}
          \STATE Set $\mathbf x^*={\rm sgn}(\mathbf v) \odot \min( |\mathbf  v|, t^* )$
    \ELSE
          \STATE Compute $c^*$, the unique root of $\phi(c)$, via
                 bisection in the interval $[\underline c, \overline
                 c]$ (Theorem~\ref{theorem:singleZF})
          \STATE Obtain $\mathbf x^*$ as the unique root of $\varphi_{c^*}^{\mathbf v}(\cdot)$
    \ENDIF
\end{algorithmic}
\end{algorithm}

\section{Experiments}\label{s:experiment}

We have conducted experiments to evaluate the efficiency of the
proposed algorithm using both synthetic and real-world data. We set
the regularization parameter as $\lambda= r \times
\lambda_{\max}^q$, where $0 < r \leq 1$ is the ratio, and
$\lambda_{\max}^q$ is the maximal value above which the
$\ell_1/\ell_q$-norm regularized problem
(\ref{eq:optimization:problem}) obtains a zero solution (see
Theorem~\ref{theorem:solution}). We try the following values for
$q$: $1.25, 1.5, 1.75, 2, 2.33, 3, 5$, and $\infty$. The source
codes, included in the SLEP package~\cite{Liu:2009:SLEP:manual}, are
available
online\footnote{\url{http://www.public.asu.edu/~jye02/Software/SLEP/}}.

\subsection{Simulation Studies}

We use the synthetic data to study the effectiveness of the
$\ell_1/\ell_q$-norm regularization for reconstructing the jointly
sparse matrix under different values of $q>1$. Let $A \in
\mathbb{R}^{m \times d}$ be a measurement matrix with entries being
generated randomly from the standard normal distribution, $X^* \in
\mathbb{R}^{d \times k}$ be the jointly sparse matrix with the first
$\tilde d < d$ rows being nonzero and the remaining rows exactly
zero, $Y = A X^* + Z$ be the response matrix, and $Z \in
\mathbb{R}^{m \times k}$ is the noise matrix whose entries are drawn
randomly from the normal distribution with mean zero and standard
deviation $\sigma=0.1$. We treat each row of $X^*$ as a group, and
estimate $X^*$ from $A$ and $Y$ by solving the following
$\ell_1/\ell_q$-norm regularized problem: $$
    X= \arg \min_{W} \frac{1}{2}\|AW-Y\|_F^2 + \lambda \sum_{i=1}^d \|W^i\|_q,$$
where $W^i$ denotes the $i$-th row of $W$. We set $m=100$, $d=200$,
and $\tilde d=k=50$. We try two different settings for $X^*$, by
drawing its nonzero entries randomly from 1) the uniform
distribution in the interval $[0,1]$ and 2) the standard normal
distribution.

\begin{figure}
  \centering
  \includegraphics[width=2.1in]{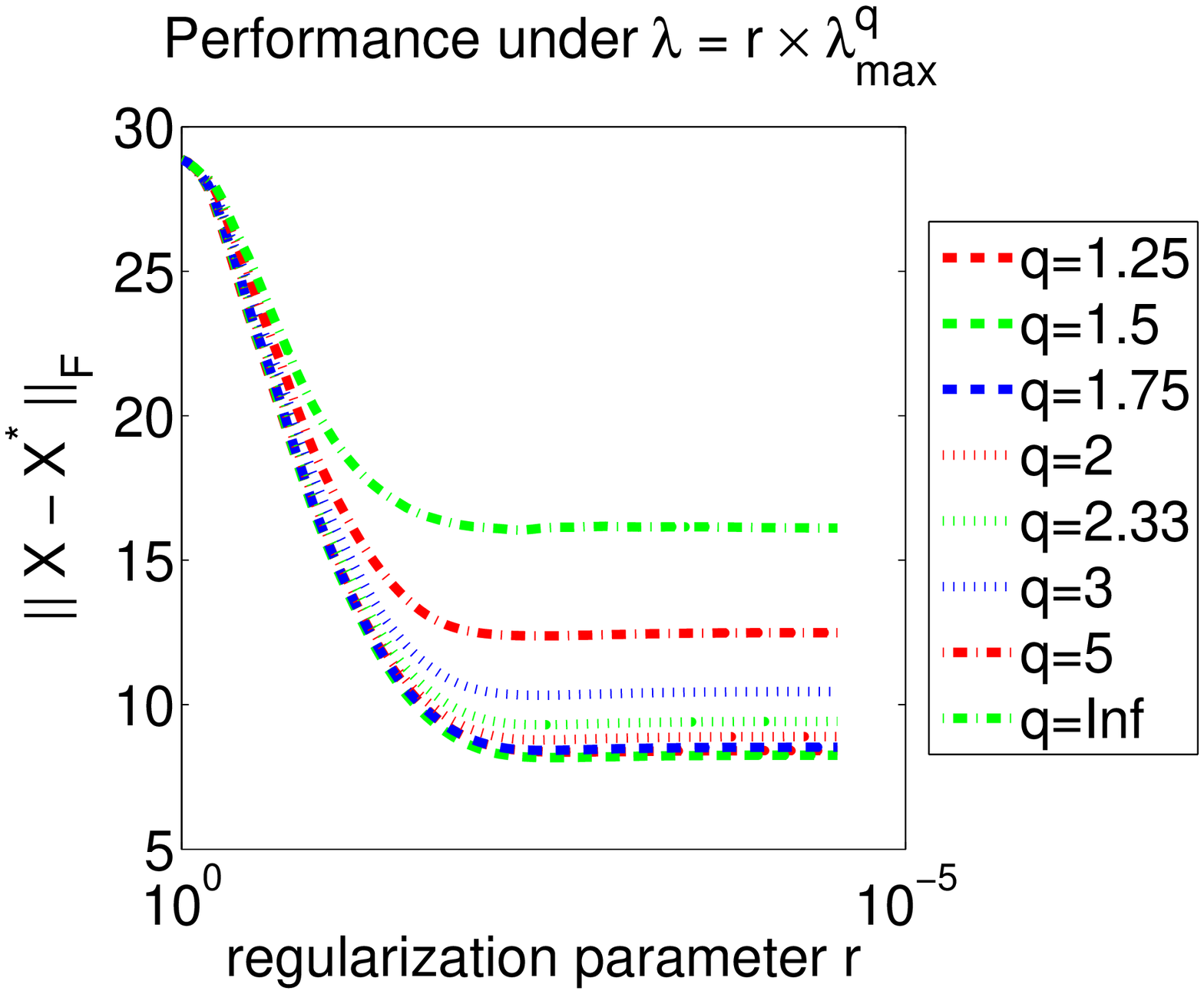}
  \hspace{0.2in}
  \includegraphics[width=2.1in]{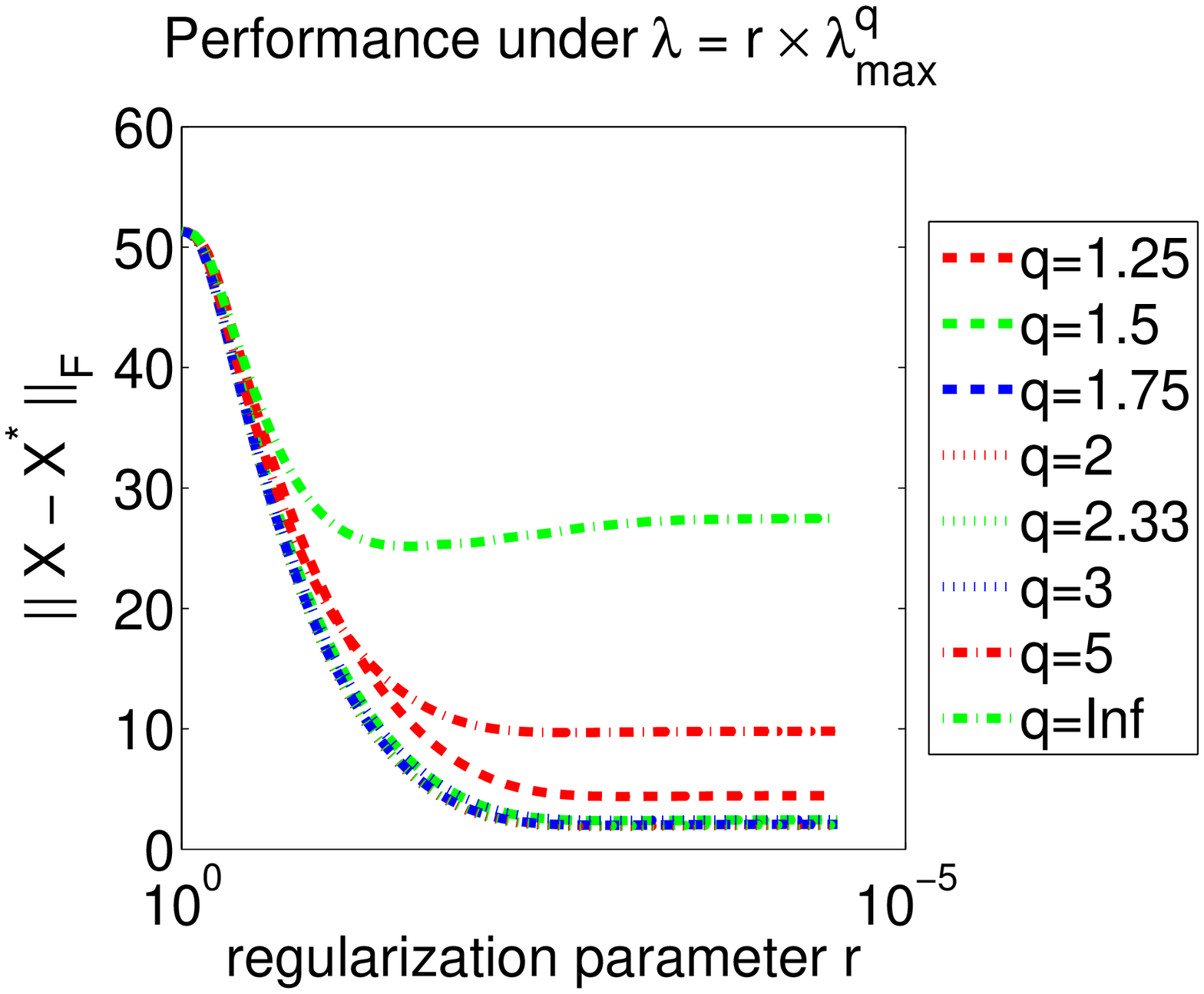}\\

  \includegraphics[width=2.1in]{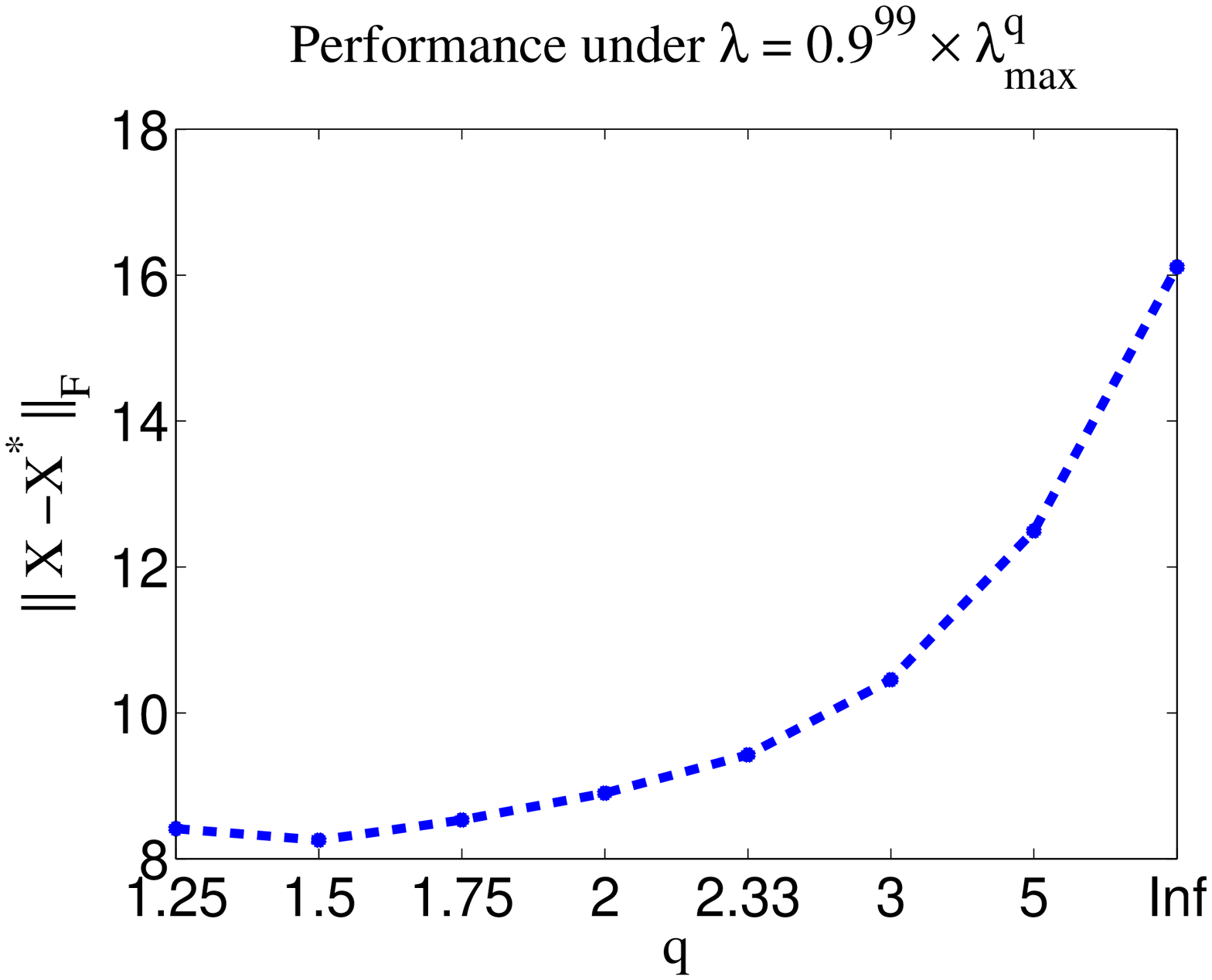}
  \hspace{0.2in}
  \includegraphics[width=2.1in]{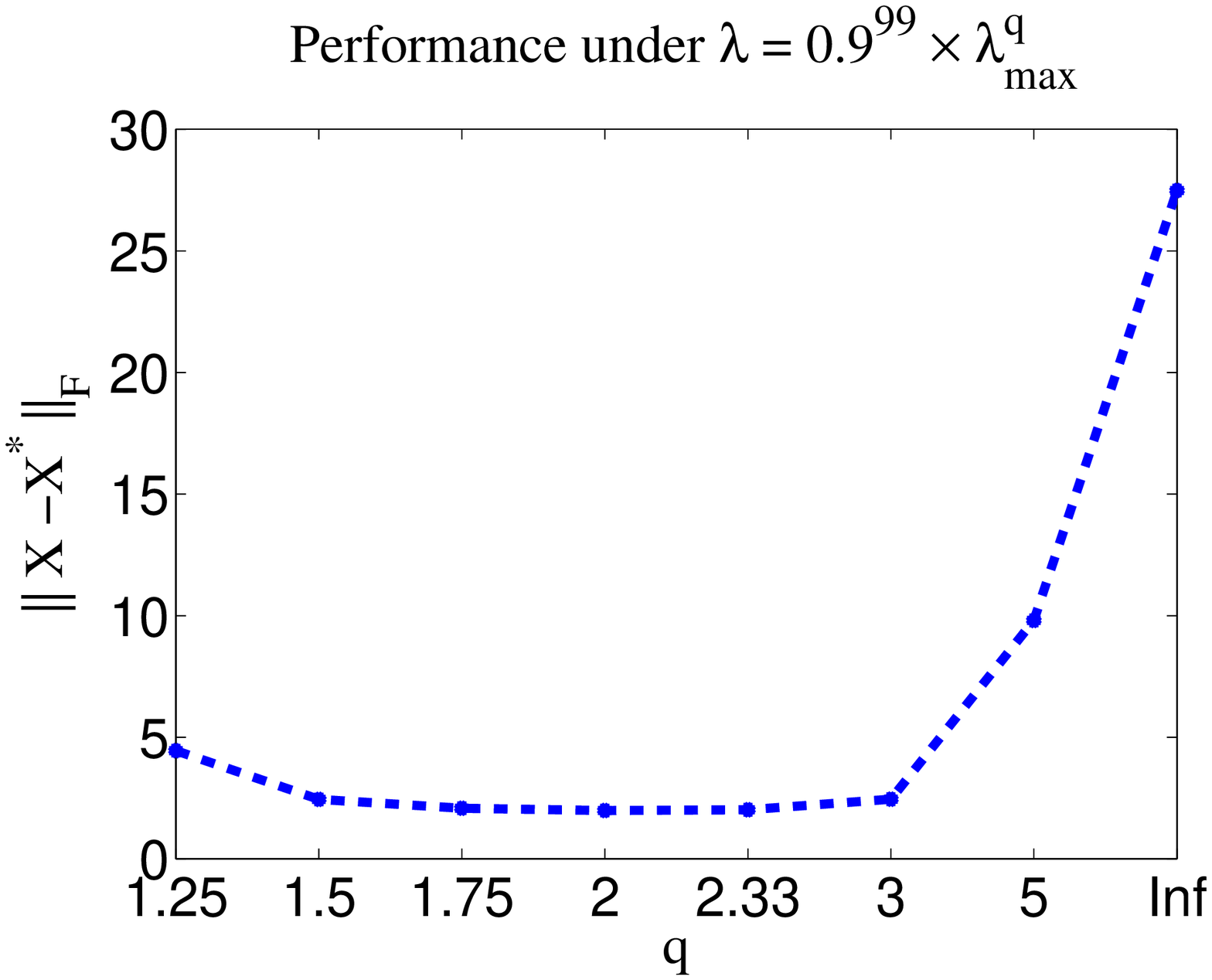}\\

  \includegraphics[width=2.1in]{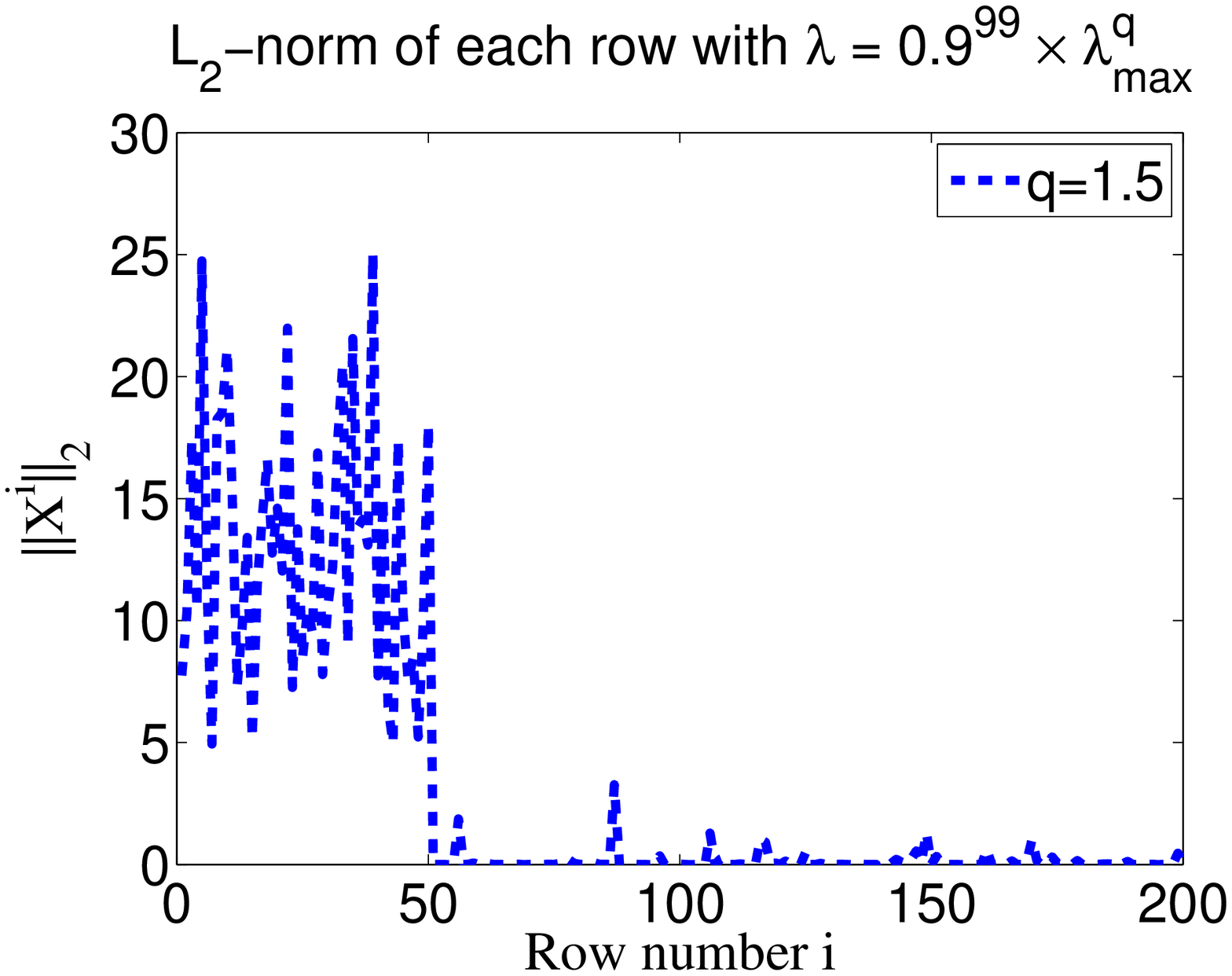}
  \hspace{0.2in}
  \includegraphics[width=2.1in]{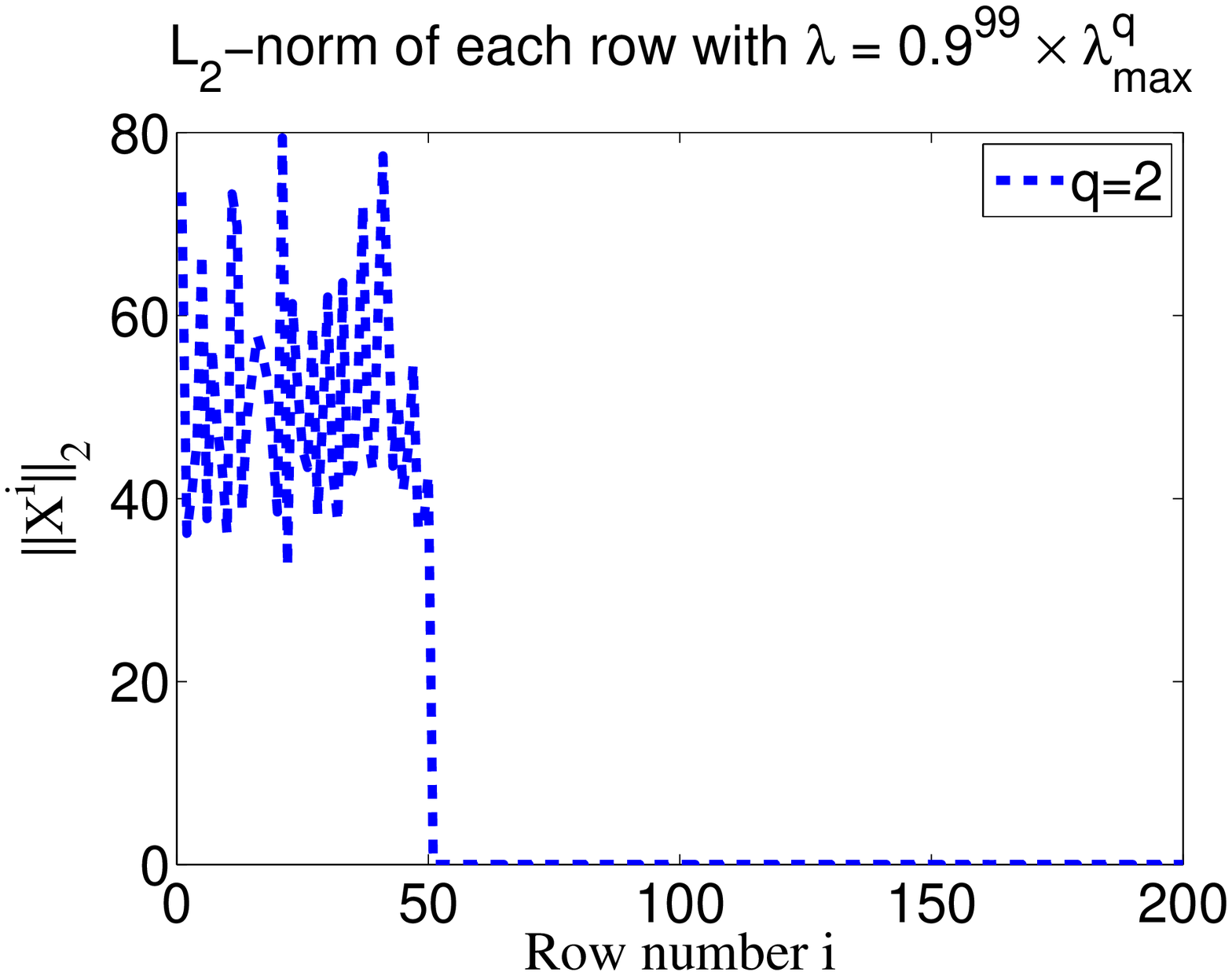}\\
\caption{ {\small Performance of the $\ell_1/\ell_q$-norm
regularization for reconstructing the jointly sparse $X^*$. The
nonzero entries of $X^*$ are  drawn randomly from the uniform
distribution for the plots in the first row, and from the normal
distribution for the plots in the second row. Plots in the first two
rows show $\|X-X^*\|_F$, the Frobenius norm difference between the
solution and the truth; and plots in the third row show the
$\ell_2$-norm of each row of the solution
$X$.}}\label{fig:synthetic}
\end{figure}

We compute the solutions corresponding to a sequence of decreasing
values of $\lambda= r \times \lambda_{\max}^q$, where $r=0.9^{i-1}$,
for $i=1, 2, \ldots, 100$. In addition, we use the solution
corresponding to the $0.9^i \times \lambda_{\max}^q$ as the ``warm"
start for $0.9^{i+1} \times \lambda_{\max}^q$. We report the results
in Figure~\ref{fig:synthetic}, from which we can observe: 1) the
distance between the solution $X$ and the truth $X^*$ usually
decreases with decreasing values of $\lambda$; 2) for the uniform
distribution (see the plots in the first row), $q=1.5$ performs the
best; 3) for the normal distribution (see the plots in the second
row), $q=1.5, 1.75, 2$ and 3 achieve comparable performance and
perform better than $q=1.25$, 5 and $\infty$; 4) with a properly
chosen threshold, the support of $X^*$ can be exactly recovered by
the $\ell_1/\ell_q$-norm regularization with an appropriate value of
$q$, e.g., $q=1.5$ for the uniform distribution, and $q=2$ for the
normal distribution; and 5) the recovery of $X^*$ with nonzero
entries drawn from the normal distribution is easier than that with
entries generated from the uniform distribution.

The existing theoretical
results~\cite{Liu:han:2009:report,Negahban:2009} can not tell which
$q$ is the best; and we believe that the optimal $q$ depends on the
distribution of $X^*$, as indicated from the above results.
Therefore, it is necessary to conduct the distribution-specific
theoretical studies (note that the previous studies usually make no
assumption on $X^*$). 
The proposed GLEP$_{1q}$ algorithm shall help verify the theoretical
results to be established.

\subsection{Performance on the Letter Data Set}

We apply the proposed GLEP$_{1q}$ algorithm for multi-task learning
on the Letter data set~\cite{Obozinski:2007}, which consists of
45,679 samples from 8 default tasks of two-class classification
problems for the handwritten letters: c/e, g/y, m/n, a/g, i/j, a/o,
f/t, h/n. The writings were collected from over 180 different
writers, with the letters being represented by $8\times 16$ binary
pixel images. We use the least squares loss for $l(\cdot)$.

\begin{figure}[h]
  \centering
  \includegraphics[width=2.1in]{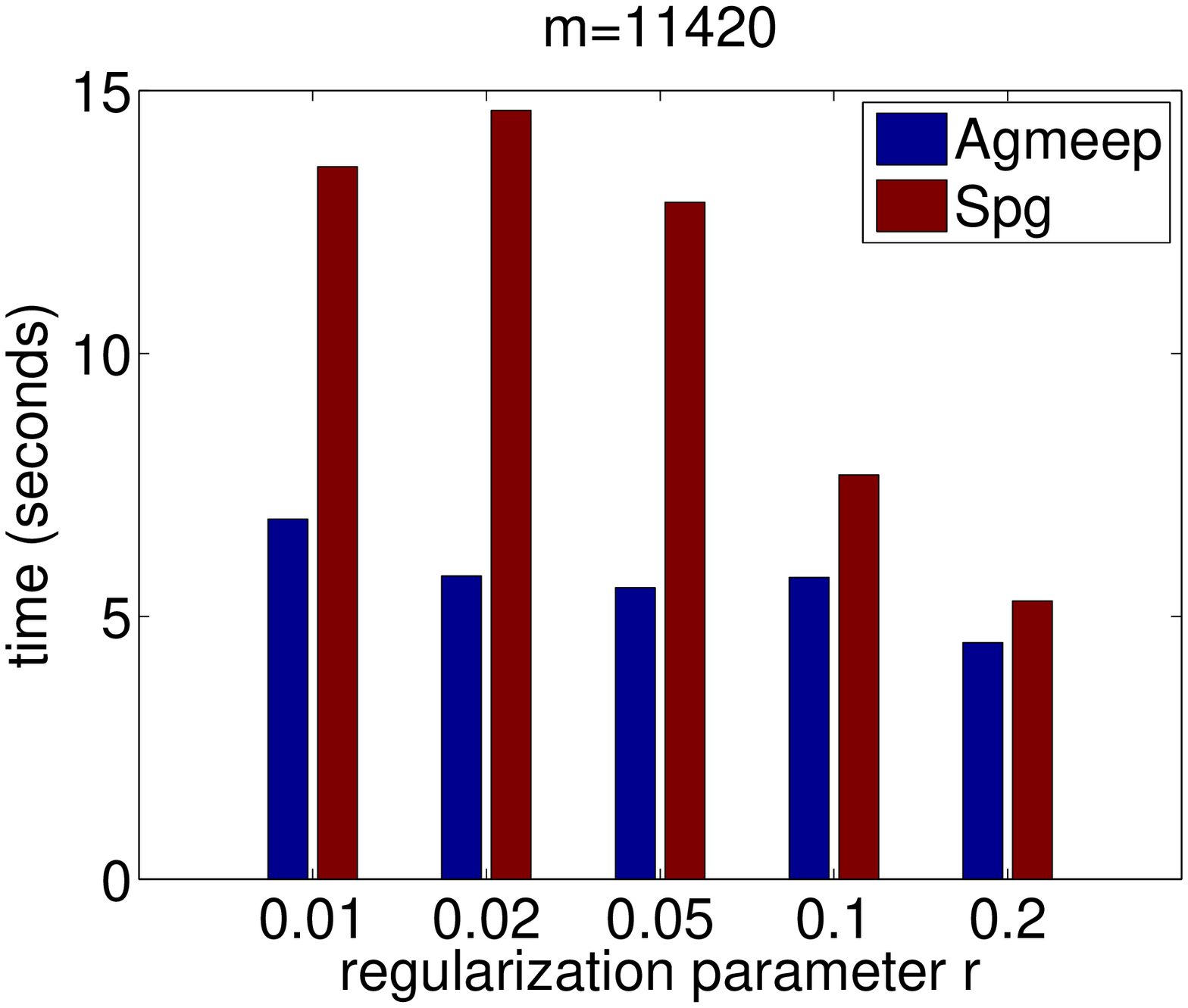}
  \includegraphics[width=2.1in]{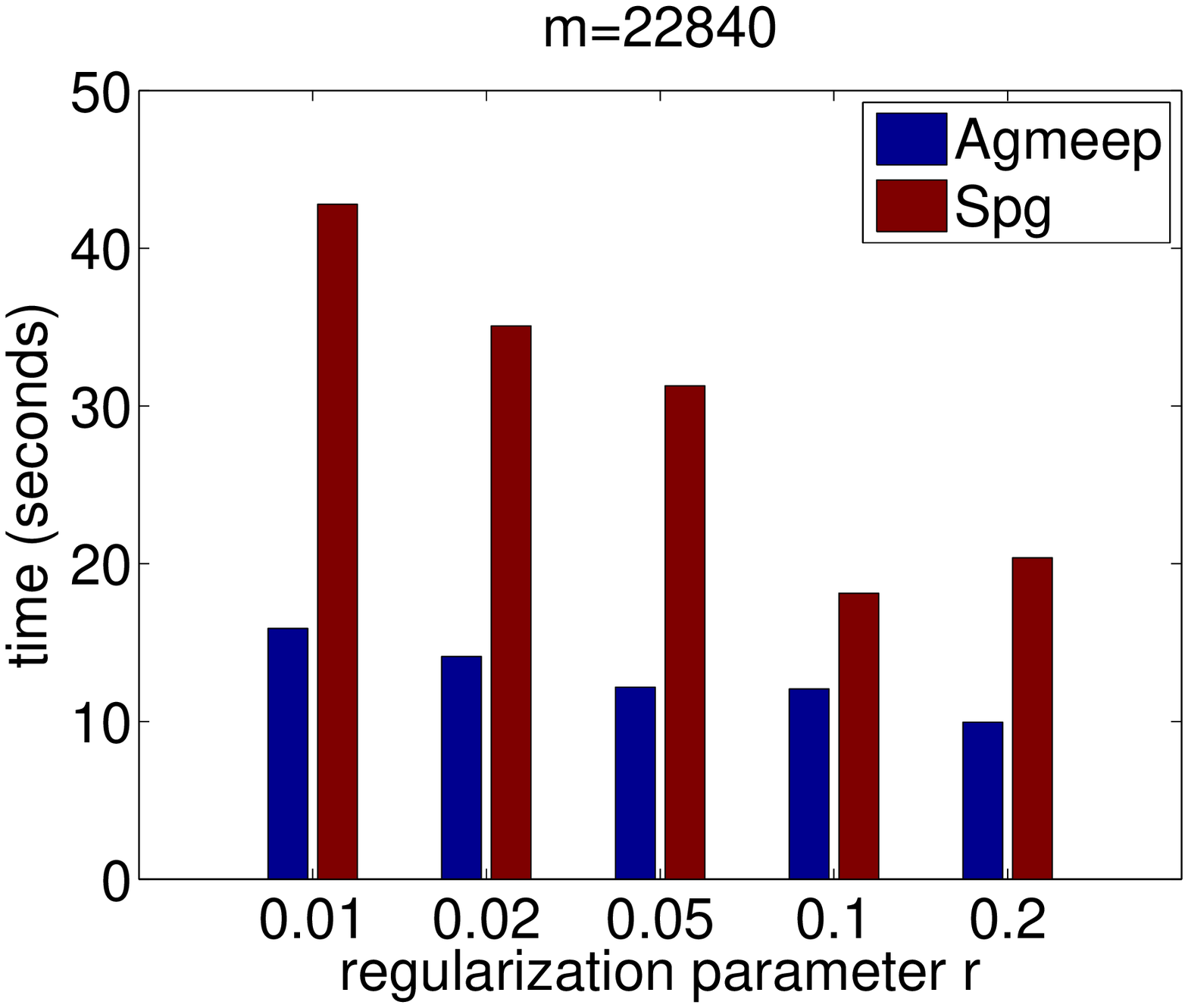}\\
  \includegraphics[width=2.1in]{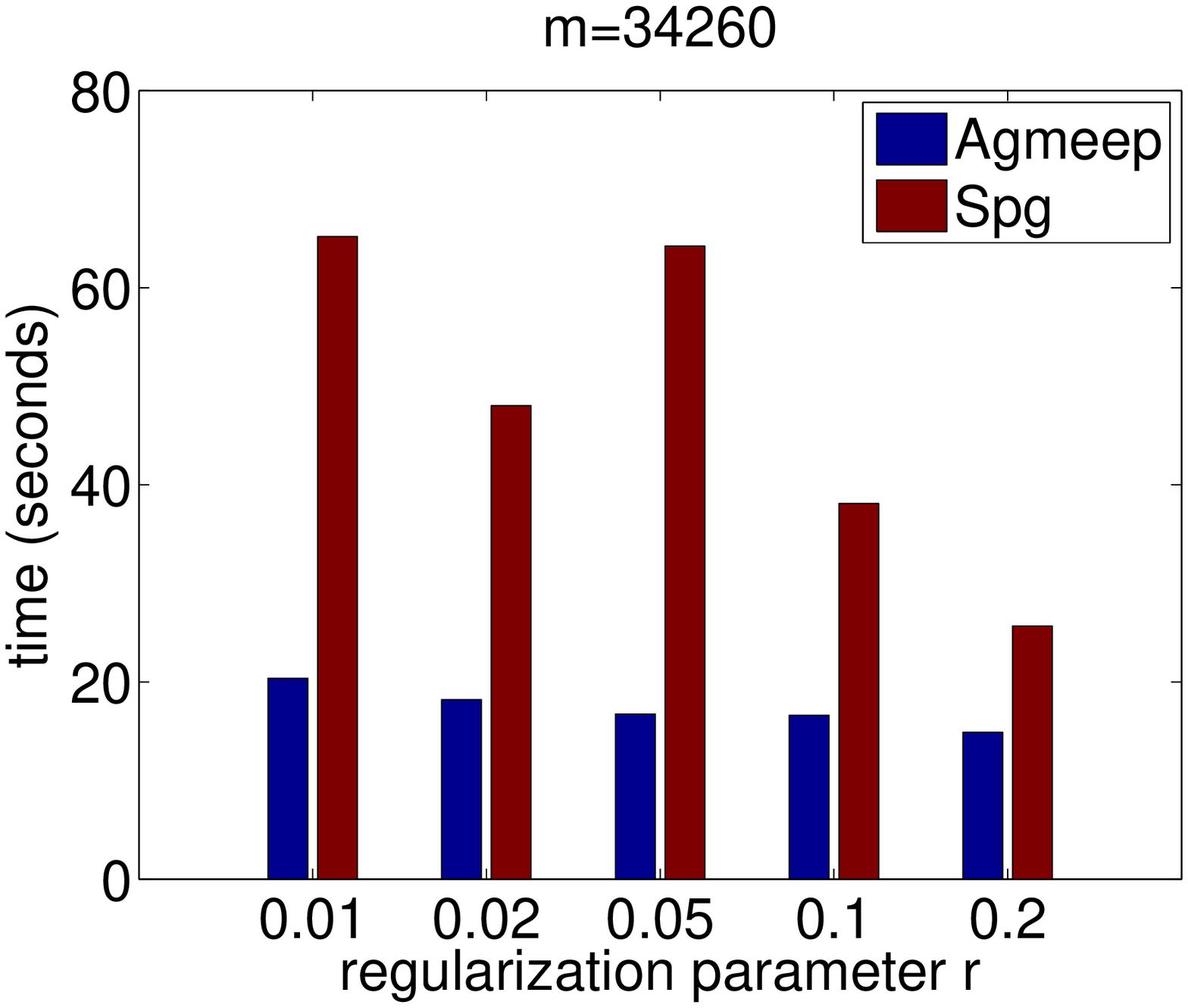}
  \includegraphics[width=2.1in]{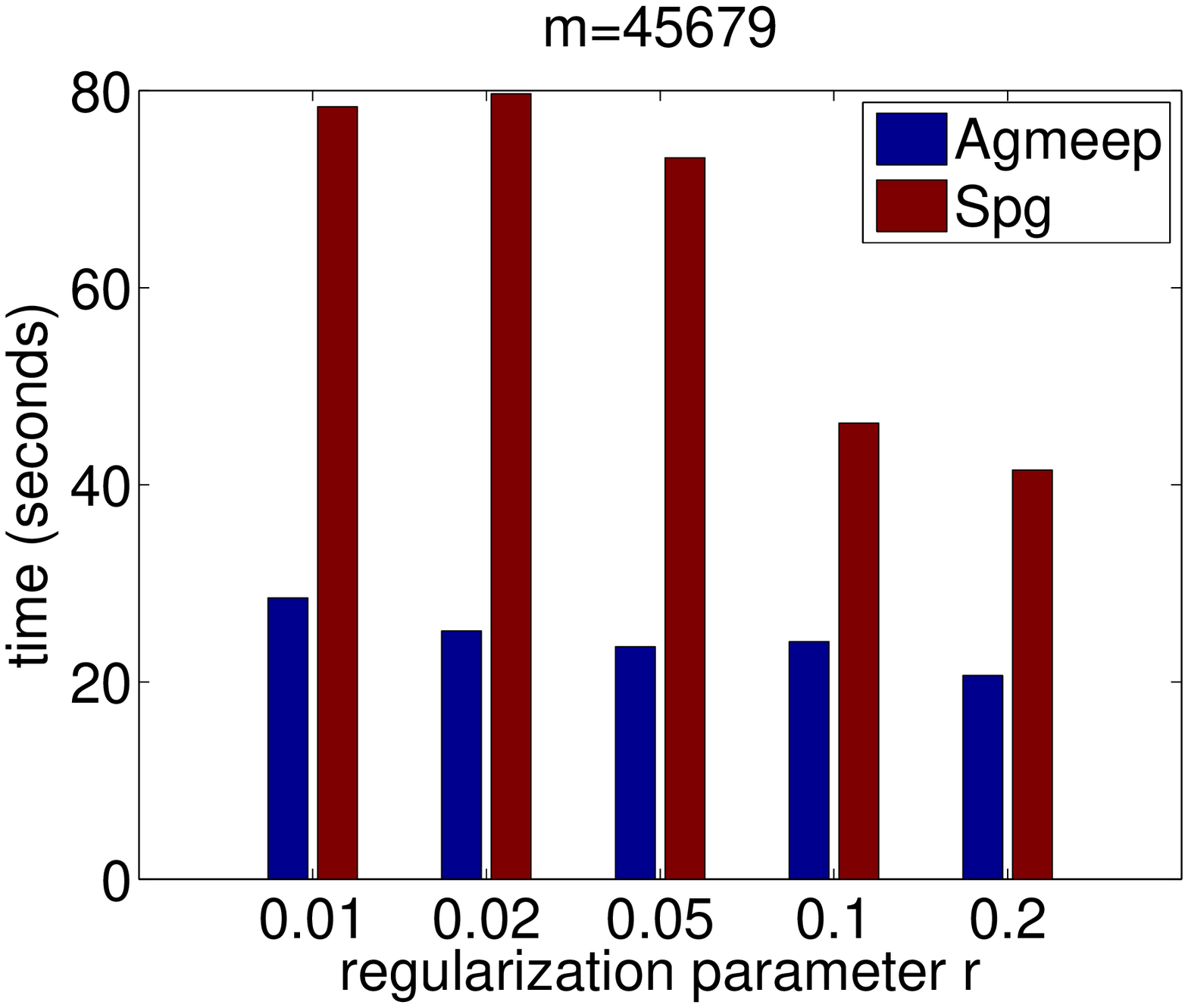}
  \caption{{\small Computational time (seconds) comparison between GLEP$_{1q}$ ($q=2$) and Spg under different values
  of $\lambda=r \times \lambda_{\max}^q$ and $m$.}}\label{fig:comp}
\end{figure}

\subsubsection{Efficiency Comparison with Spg} We compare GLEP$_{1q}$
with the Spg algorithm proposed in~\cite{BergFriedlander:2008}. Spg
is a specialized solver for the $\ell_1/\ell_2$-ball constrained
optimization problem, and has been shown to outperform existing
algorithms based on blockwise coordinate descent and projected
gradient. In Figure~\ref{fig:comp}, we report the computational time
under different values of $m$ (the number of samples) and $\lambda=r
\times \lambda_{\max}^q$ ($q=2$). It is clear from the plots that
GLEP$_{1q}$ is much more efficient than Spg, which may attribute to:
1) GLEP$_{1q}$ has a better convergence rate than Spg; and 2) when
$q=2$, the EP$_{1q}$ in GLEP$_{1q}$ can be computed analytically (see
Remark~\ref{remark:q=2}), while this is not the case in Spg.

\subsubsection{Efficiency under Different Values of $q$ } 
\begin{figure}
  \centering
  \includegraphics[width=2.1in]{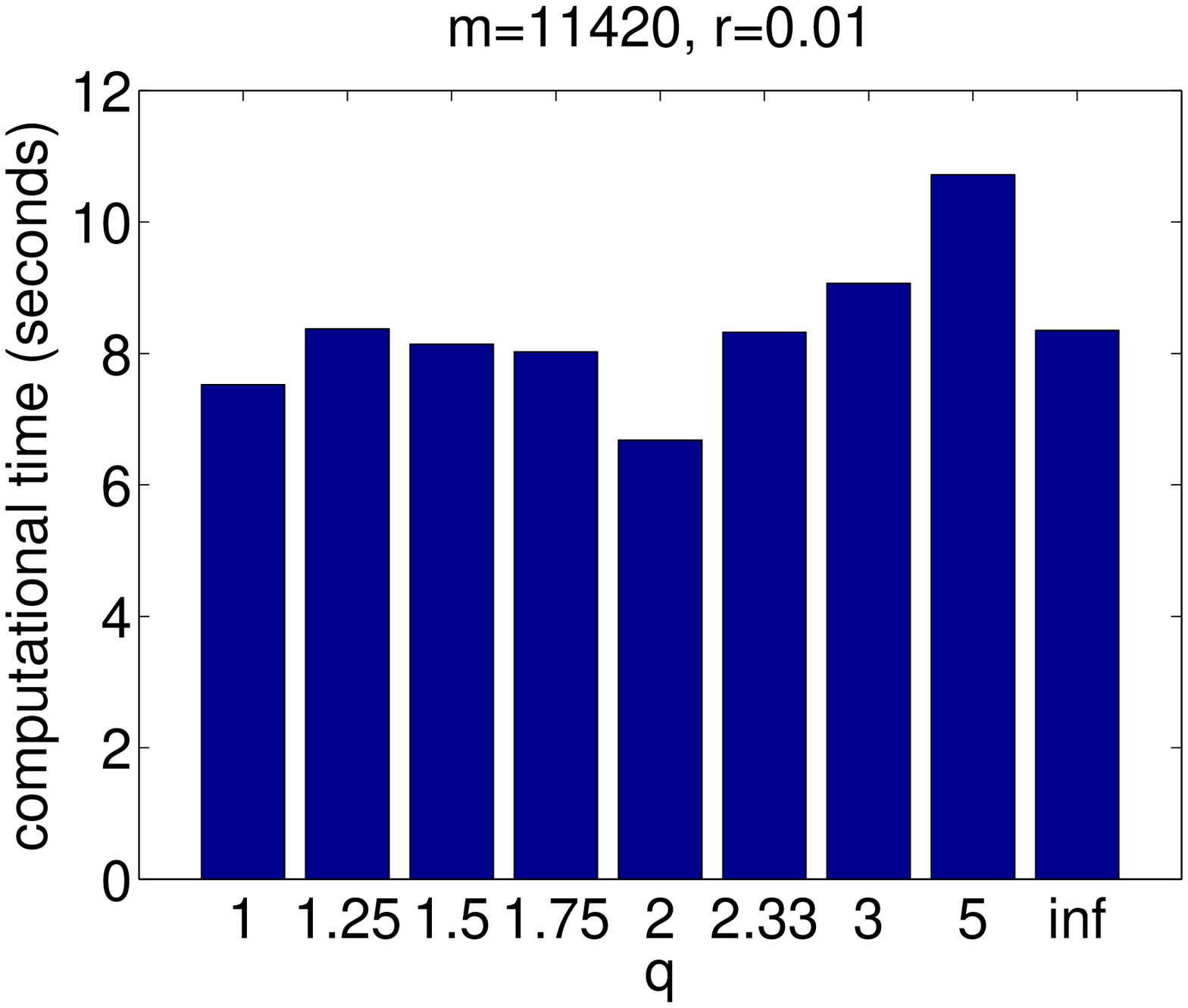}
  \includegraphics[width=2.1in]{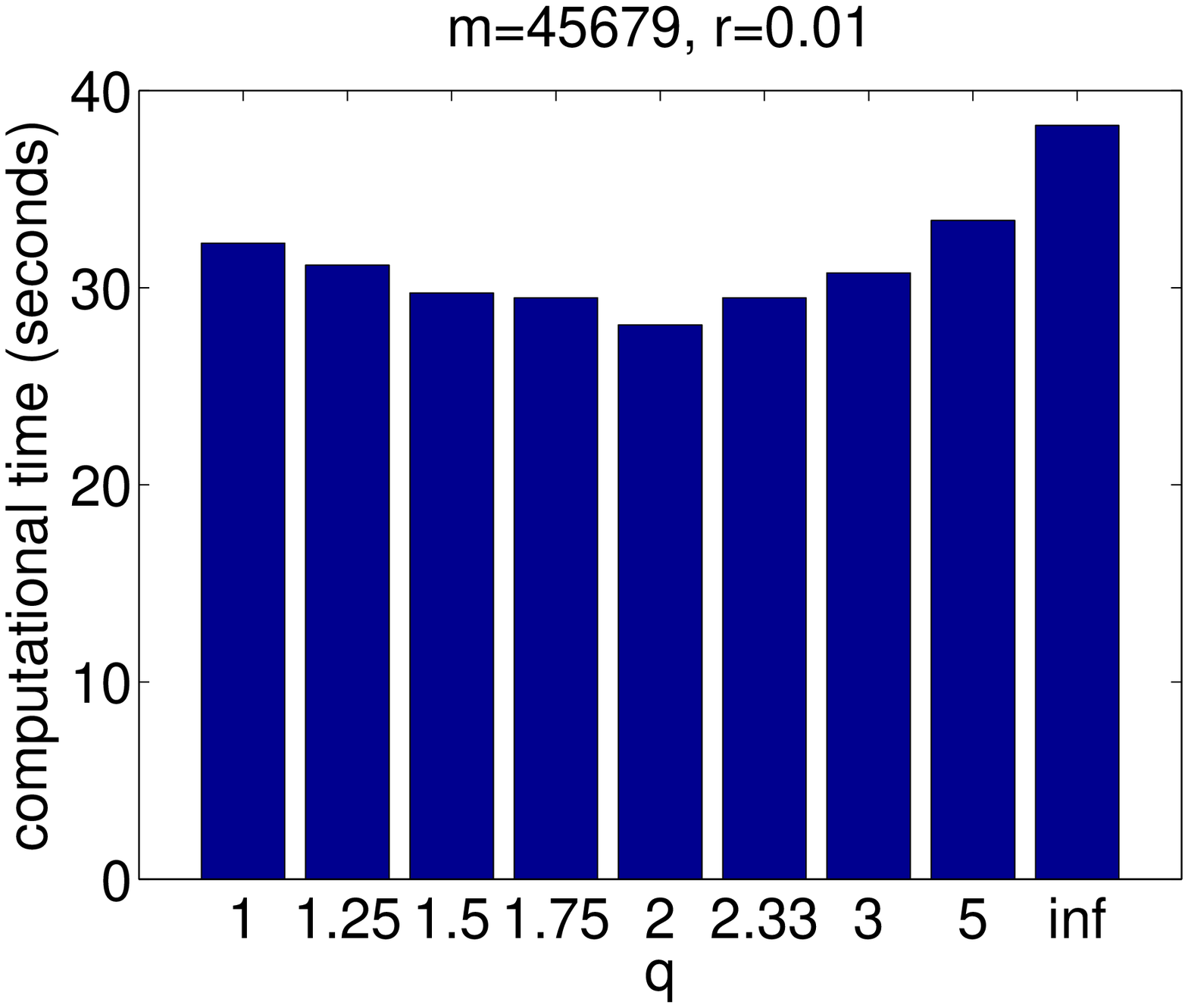}\\
  \includegraphics[width=2.1in]{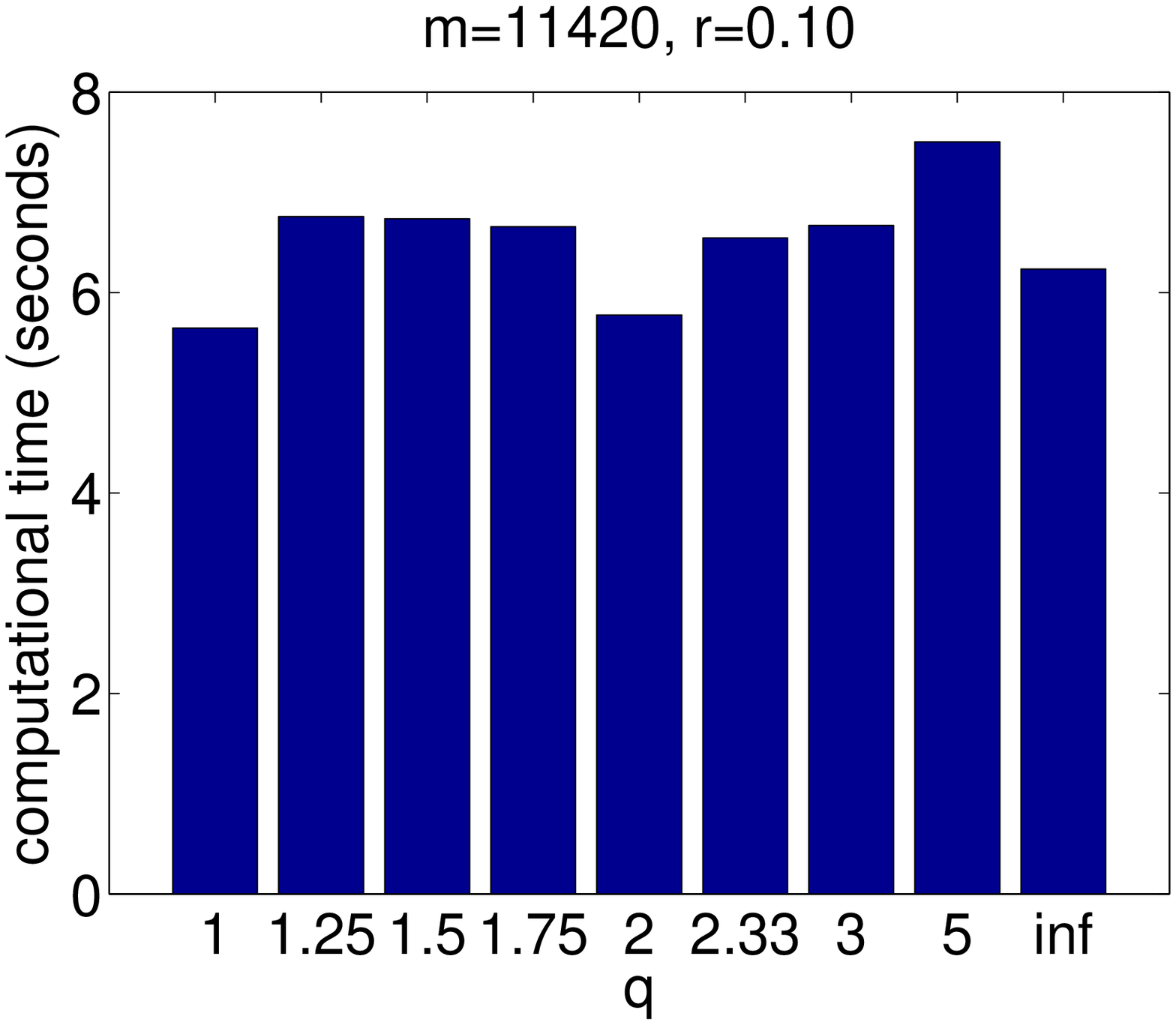}
  \includegraphics[width=2.1in]{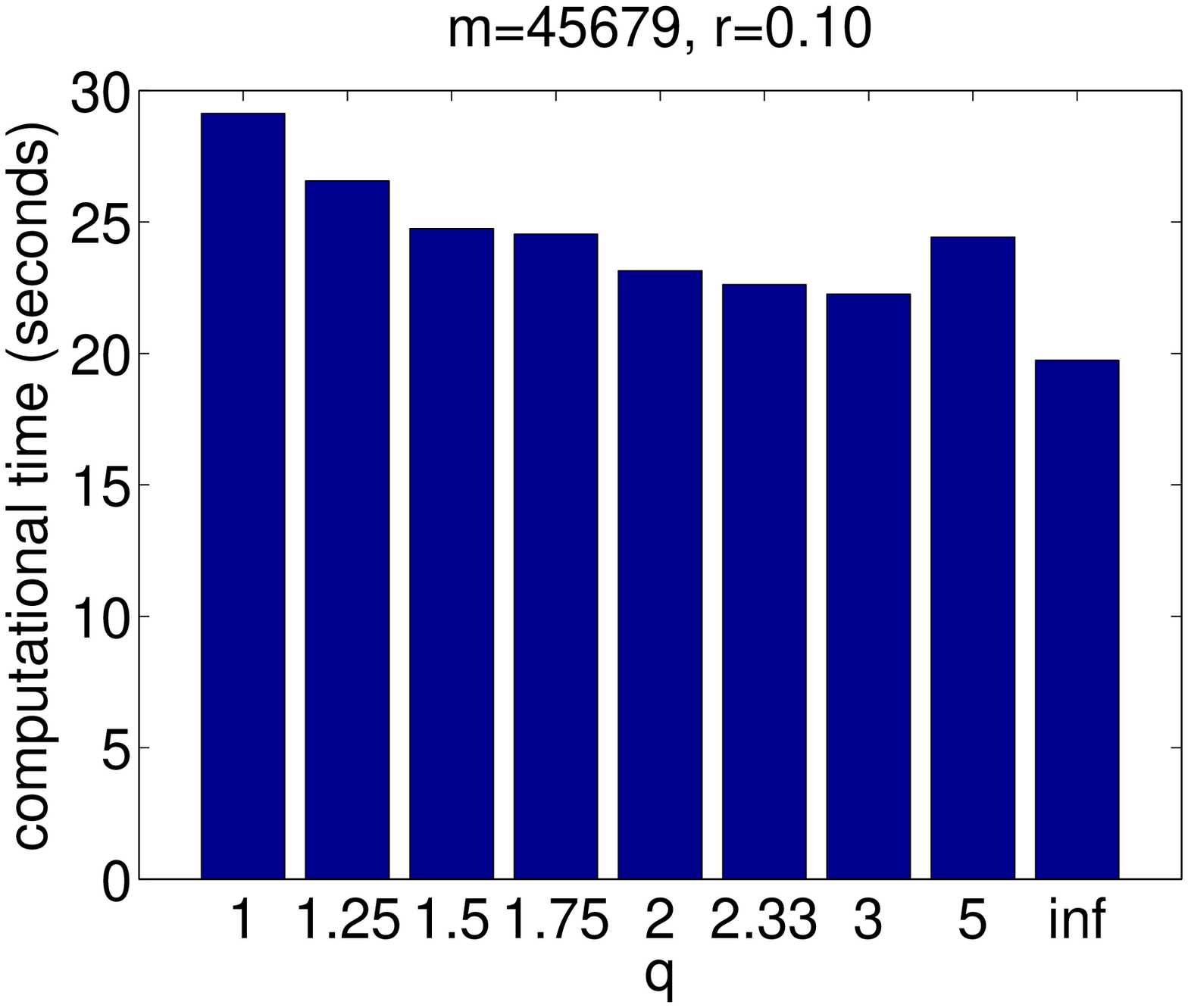}\\
  \caption{{\small Computation time (seconds) of GLEP$_{1q}$ under different values of
  $m$, $q$ and $r$.}} \label{fig:time}
\end{figure}

We
report the computational time (seconds) of GLEP$_{1q}$ under
different values of $q$, $\lambda=r \times \lambda_{\max}^q$ and $m$
(the number of samples) in Figure~\ref{fig:time}. We can observe from
this figure that the computational time of GLEP$_{1q}$ under
different values of $q$ (for fixed $r$ and $m$) is comparable.
Together with the result on the comparison with Spg for $q=2$, this
experiment shows the promise of GLEP$_{1q}$ for solving large-scale
problems for any $q \geq 1$.


\begin{figure}
  \centering
  \includegraphics[width=2.1in]{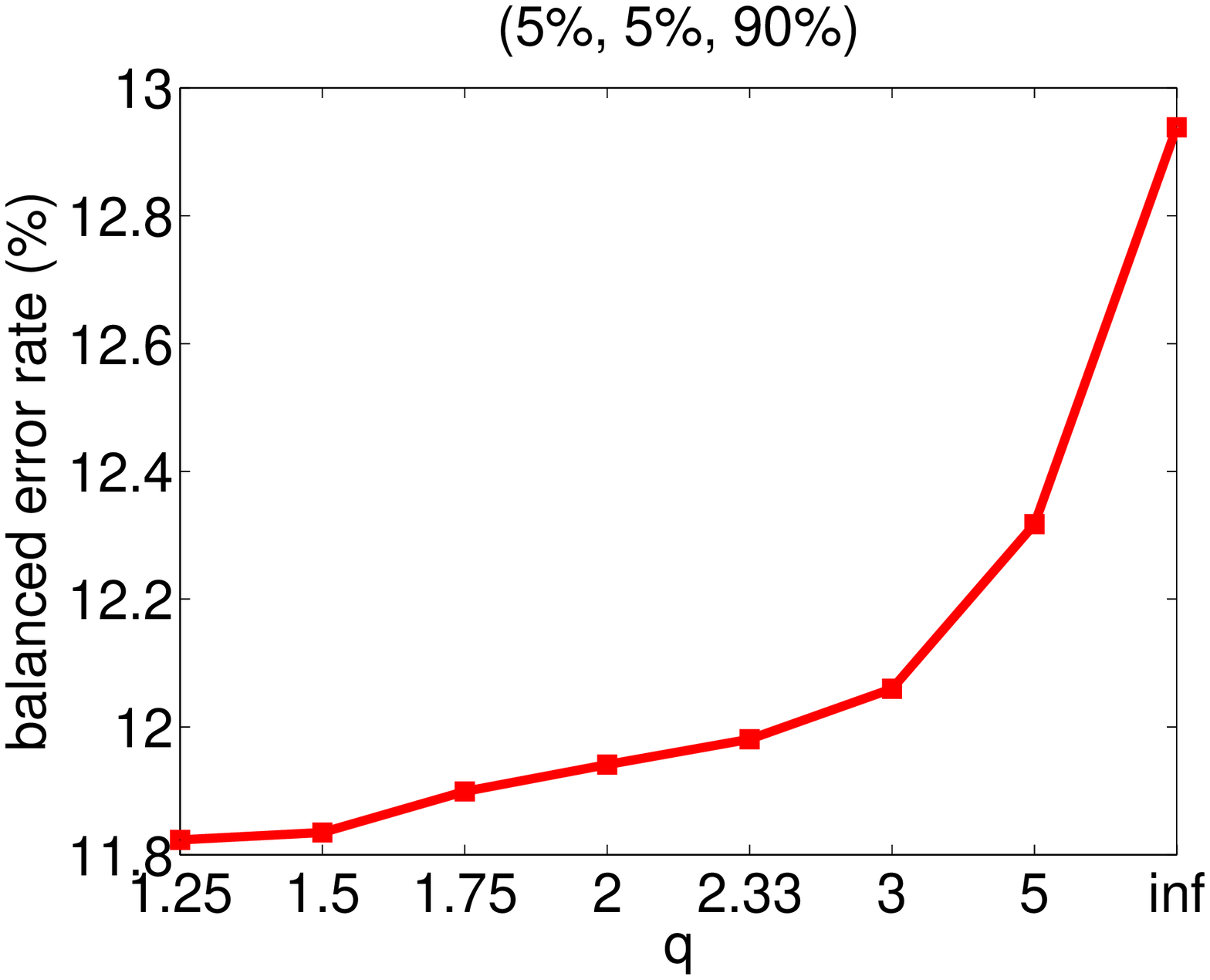}
  \includegraphics[width=2.1in]{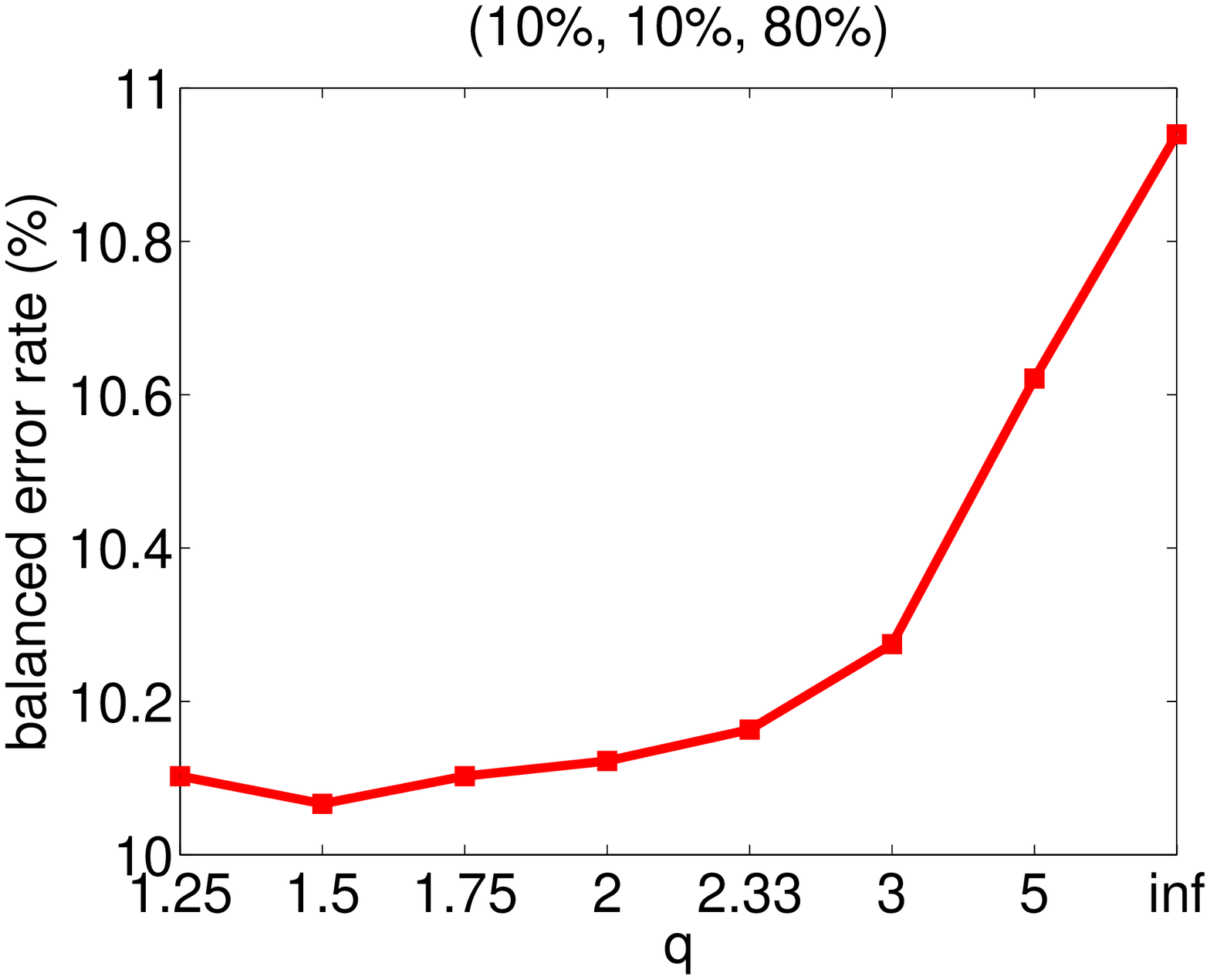}\\
  \caption{{\small
The balanced error rate achieved by the $\ell_1/\ell_q$
regularization under different values of $q$. The title of each plot
indicates the percentages of samples used for training, validation,
and testing.}} \label{fig:letter:performance}
\end{figure}

\subsubsection{Performance under Different Values of $q$ } We
randomly divide the Letter data into three non-overlapping sets:
training, validation, and testing. We train the model using the
training set, and tune the regularization parameter $\lambda=r \times
\lambda_{\max}^q$ on the validation set, where $r$ is chosen from $\{
10^{-1}, 5\times 10^{-2}, 2 \times 10^{-2}, 1 \times 10^{-2}, 5
\times 10^{-3}, 2 \times 10^{-3}, 1 \times 10^{-3}\}$. On the testing
set, we compute the balanced error rate~\cite{Guyon:2004}. We report
the results averaged over 10 runs in
Figure~\ref{fig:letter:performance}. The title of each plot indicates
the percentages of samples used for training, validation, and
testing. The results show that, on this data set,  a smaller value of
$q$ achieves better performance.

\section{Conclusion}\label{s:conclusion}

In this paper, we propose the GLEP$_{1q}$ algorithm for solving the
$\ell_1/\ell_q$-norm regularized problem, for any $q \geq 1$. The
main technical contribution of this paper is the efficient algorithm
for the $\ell_1/\ell_q$-norm regularized Euclidean projection
(EP$_{1q}$), which is a key building block of GLEP$_{1q}$.
Specifically, we analyze the key theoretical properties of the
solution of EP$_{1q}$, based on which we develop an efficient
algorithm for EP$_{1q}$ by solving two zero finding problems. Our
analysis also reveals why EP$_{1q}$ for the general $q$ is
significantly more challenging than the special cases such as $q=2$.


In this paper, we focus on the efficient implementation of the
$\ell_1/\ell_q$-regularized problem. We plan to study the
effectiveness of the $\ell_1/\ell_q$ regularization under different
values of $q$ for real-world applications in computer vision and
bioinformatics. We also plan to conduct the
distribution-specific~\cite{Eldar:2010} theoretical studies for
different values of $q$.

%

\end{document}